\documentclass[10pt,twocolumn,letterpaper]{article}

\usepackage{iccv}
\usepackage{times}
\usepackage{epsfig}
\usepackage{graphicx}
\usepackage{amsmath}
\usepackage{amssymb}

\usepackage{booktabs}
\usepackage{algorithm}
\usepackage{algorithmic}
\usepackage{amsfonts}
\usepackage{subfig}
\usepackage[export]{adjustbox}

\newlength{\tempdim}

\iccvfinalcopy 

\def\iccvPaperID{****} 

\ificcvfinal\fi

\begin{document}

\title{Manga Rescreening with Interpretable Screentone Representation}

\author{
\begin{tabular}{ccccc}
Minshan Xie\textsuperscript{\rm 1} &
Chengze Li\textsuperscript{\rm 2} &
Tien-Tsin Wong\textsuperscript{\rm 1}\thanks{Corresponding author.}
 \end{tabular}\\
\begin{tabular}{cc}
\textsuperscript{\rm 1} The Chinese University of Hong Kong & 
\textsuperscript{\rm 2} Caritas Institute of Higher Education\\
 \end{tabular}\\
\begin{tabular}{cc}
{\tt\small \{msxie, ttwong\}@cse.cuhk.edu.hk} &
{\tt\small czli@cihe.edu.hk}
 \end{tabular}
}

\maketitle
\ificcvfinal\thispagestyle{empty}\fi

\begin{abstract}
The process of adapting or repurposing manga pages is a time-consuming task that requires manga artists to manually work on every single screentoned region and apply new patterns to create novel screentones across multiple panels. To address this issue, we propose an automatic manga rescreening pipeline that aims to minimize the human effort involved in manga adaptation. Our pipeline automatically recognizes screentone regions and generates novel screentones with newly specified characteristics (e.g., intensity or type). 
Existing manga generation methods have limitations in understanding and synthesizing complex tone- or intensity-varying regions. To overcome these limitations, we propose a novel interpretable representation of screentones that disentangles their intensity and type features, enabling better recognition and synthesis of screentones. This interpretable screentone representation reduces ambiguity in recognizing intensity-varying regions and provides fine-grained controls during screentone synthesis by decoupling and anchoring the type or the intensity feature.
Our proposed method is demonstrated to be effective and convenient through various experiments, showcasing the superiority of the newly proposed pipeline with the interpretable screentone representations. 

\end{abstract}


\section{Introduction}
\label{sec:intro}

\begin{figure}[t!]
    \centering
    \subfloat[Manga]{
    \adjincludegraphics[clip,trim=0 0 0 {.1\width},width=.24\linewidth]{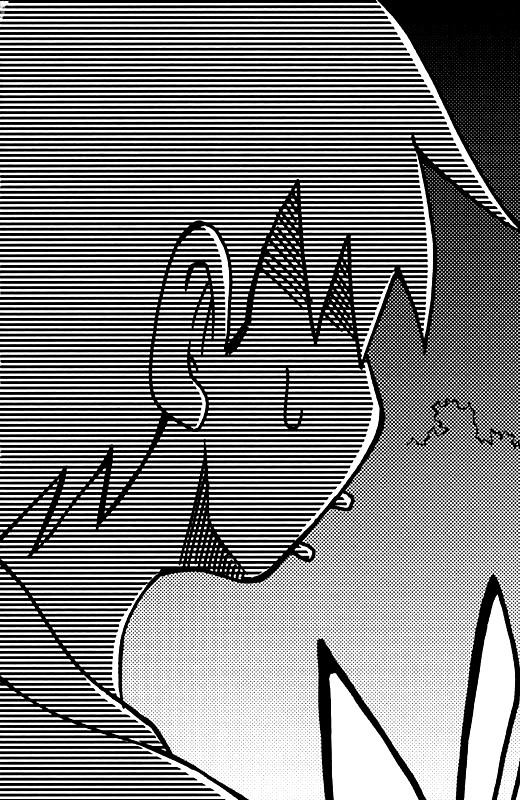}}
    \subfloat[\centering Segmentation based on type]{
    \adjincludegraphics[clip,trim=0 0 0 {.1\width},width=.24\linewidth]{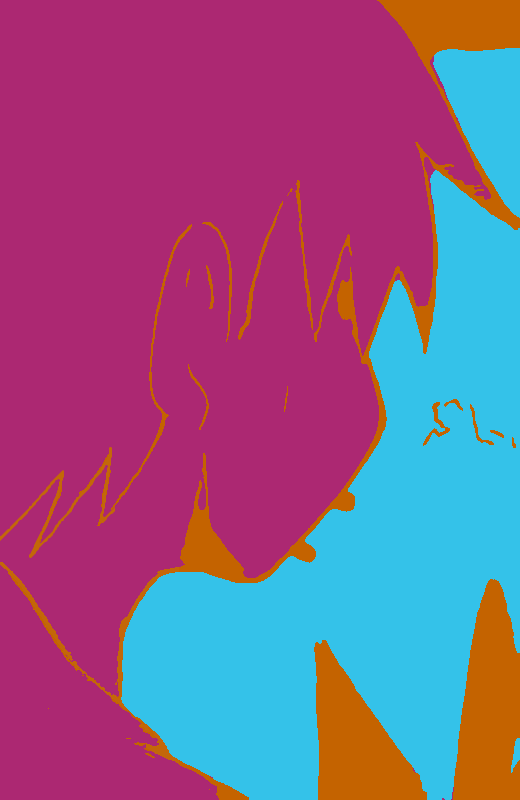}}
    \subfloat[\centering Type alternation]{
    \adjincludegraphics[clip,trim=0 0 0 {.1\width},width=.24\linewidth]{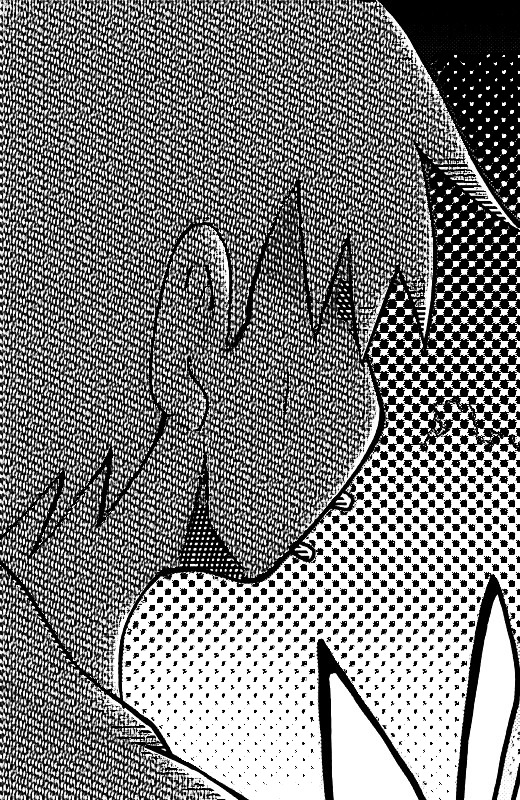}}
    \subfloat[\centering Intensity alternation]{
    \adjincludegraphics[clip,trim=0 0 0 {.1\width},width=.24\linewidth]{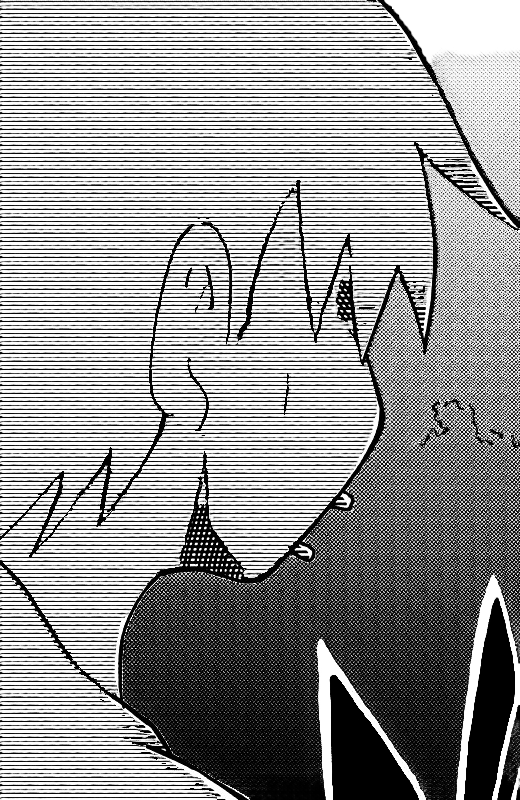}}
    \caption{Our method can identify manga screentone regions and edit them by either intensity or screentone type. }
    \vspace{-.1in}
    \label{fig:rescreening}
\end{figure}

Japanese manga is a worldwide entertainment art form. It is unique in its region-filling with specially designed bitonal patterns, called \textit{screentones}, to enrich its visual contents. During manga production, artists will carefully select these screentones considering both intensity and pattern type to express various shading effects~\cite{qu2008richness}. 
The screentone intensity, similarly as the luminance channel in color images, plays the key role to render and represent shading among different regions, while the screentone type is an alternative to the chrominance channel in color images to differentiate semantics over regions. 
However, once the manga is produced, this screening process is hard to be reverted for editing or adaptation to other types of medium, such as e-readers, Webtoons, etc. Even if the extraction and editing of ink line drawings are easy, the adaptation of screentones is exceptionally challenging, as simple pixel-level transformation of screentoned area tend to break the intended visual delivery by the artists, such as the fading effects or region contrast. As a result, it usually requires a complete rescreening of the whole manga frame. In such a process, the artists have to manually identify and amend the regions for correction (Fig.~\ref{fig:rescreening}(b)). The overall process is tedious and labor-intensive, as one has to handle diverse cases of screentone variation in both screentone intensity and type. 

To save time and cost, computer techniques are considered to be employed to ease the rescreening process, but the bitonal and discrete screentones hinder both manga segmentation and screentone synthesis. 
To segment the region filled with the same screentone, recent approaches~\cite{liu2017boundary,qu2006manga} focus on the discovery of efficient texture descriptors, like Gabor filter banks~\cite{manjunath1996texture}. 
However, these methods cannot tackle intensity-varying screentones (background in Fig.~\ref{fig:rescreening}(a)) since they did not disentangle the type feature from the intensity information. 
For screentone synthesis and manga generation, several attempts are proposed to produce screened manga~\cite{jarvis1976survey,ulichney1987digital,durand2001decoupling,pang2008structure,qu2008richness,zhang2021generating}. But, they either generated more or less uniform patterns or failed to generate intensity-varying effects. 
Recently, Xie et al.~\cite{xie2020manga} proposed a representation for screentone, which supports both region discrimination and screentone synthesis. However, since intensity and type features are unrecognizable in their model, it limits the user to identifying screentone regions with varying intensity and synthesizing screentones with given intensity or type. 
As can be observed in Fig.~\ref{fig:rescreening}, manga artists commonly use the same type of screentone, sometimes with intensity variation, to fill the same semantic region, which requires segmenting regions based on screentone type. 

In this paper, we propose a framework to implement intuitive and convenient manga rescreening. The framework can recognize semantic continuous regions based on the screentone type, and manages to enable individual tuning of screentone type or intensity while maintaining the other one. 
The framework starts by learning an interpretable screentone representation with disentangled intensity and type features. The intensity feature is expected to visually conform to the tone/shading intensity of manga, and the type feature should have the same representation over the same screentone type. Such a representation allows discrimination of the same screentone type with varying intensity and also precise control on screentone generation given a specific tone intensity and type. 

To disentangle the intensity and type features, we propose to encode the latent domain with intensity as one \textbf{independent} axis. 
Yet, we observed that the diversity of screentone types is correlated with the intensity, where darker or lighter tone intensity shall gradually suppress the tone diversity. Thus, we propose to model the domain as a hypersphere space, and screentones with the same intensity are encoded as a normal distribution with standard deviation conditioned on the intensity. 
The disentangled representation greatly benefits region semantic understanding and segmentation of manga. While all existing texture descriptors fail in distinguishing a semantic consistent screentone region with varying intensities, the proposed representation manages to catch this consistency to produce better segmentations (Fig.~\ref{fig:rescreening}(b)). 
With manga regions segmented, our framework can further synthesize screentones with any specified effects by editing the intensity or type features. Our method can generate various screentones preserving the original intensity variation (Fig.~\ref{fig:rescreening}(c)) or the original screentone types (Fig.~\ref{fig:rescreening}(d)).

To validate the effectiveness of our method, we apply our method in various real-world cases and receive convincing results. We conclude our contributions as follows.
\begin{itemize}
    \item We propose a practical manga rescreening method with an interpretable screentone representation to enable manga segmentation and user-expected screentone generation.
    \item We disentangle the intensity information from the screentones by modeling a hypersphere space with intensity as the major axis.
\end{itemize}

\section{Related Work}

\paragraph{Manga Segmentation}
Existing manga segmentation approaches focus on the discovery of efficient texture descriptors as screentones are laid over regions rather than individual pixels. Traditional texture analysis usually analyzes the texture by first applying filtering techniques and then representing the textures with statistical models~\cite{weldon1996efficient,varma2003texture,galun2003texture,manjunath1996texture}.
Gabor Wavelet features~\cite{manjunath1996texture} has been demonstrated as an effective texture discrimination technique for screentones~\cite{qu2006manga} and utilized for measuring pattern-continuity in various manga tasks~\cite{qu2006manga,qu2008richness}. 
However, all the above texture features are windowed-based and usually fail at boundary and thin structures.  
Convolutional Neural Network (CNN) has also been shown to be suitable for texture analysis, in which trainable filter banks make an excellent tool in the analysis of repetitive texture patterns~\cite{cimpoi2014describing,cimpoi2015deep,cimpoi2016deep}.
But these methods are tailored for extracting texture features in natural images and usually fail for screened manga. 
Recently, Xie et al.~\cite{xie2020manga} attempted to identify regions in manga by mapping screentones into an interpolative representation with a Screentone Variational AutoEncoder (ScreenVAE). The model features a vast screentone encoding space, which generates similar representations for similar screentones. However, it cannot handle screentone regions of varying intensity. 
In comparison, we disentangle intensity feature from type feature and can identify the same screen type without being confused by the intensity variation.

\paragraph{Manga Generation}
Traditional manga generation shades grayscale/color images to produce screened manga through halftoning~\cite{jarvis1976survey,ulichney1987digital,pang2008structure,zhang2021generating} or hatching~\cite{winkenbach1994computer,durand2001decoupling}. But these methods reproduce the intensity with more or less uniform patterns and cannot satisfy the requirement for manga which uses a rich set of screentone types to enrich the viewing experience. Qu et al.~\cite{qu2008richness} screened color images by selecting various screentones considering tone similarity, texture similarity, and chromaticity distinguishability. However, with the limited screentone set, it may not generate screentones with user-expected intensity and types,  even generate smooth transitions with the same screentone type. 
ScreenVAE~\cite{xie2020manga} produced screened manga with an interpolative screentone representation. 
However, since intensity and type features are unrecognizable in their model, it limits the user to synthesizing new types of screentone that retain special effects. In comparison, we propose an interpretable representation disentangling intensity and type of screentones, which enables controllable generation and manipulation on both intensity and type. 

\begin{figure*}[t!]
    \centering
    \includegraphics[width=.95\linewidth]{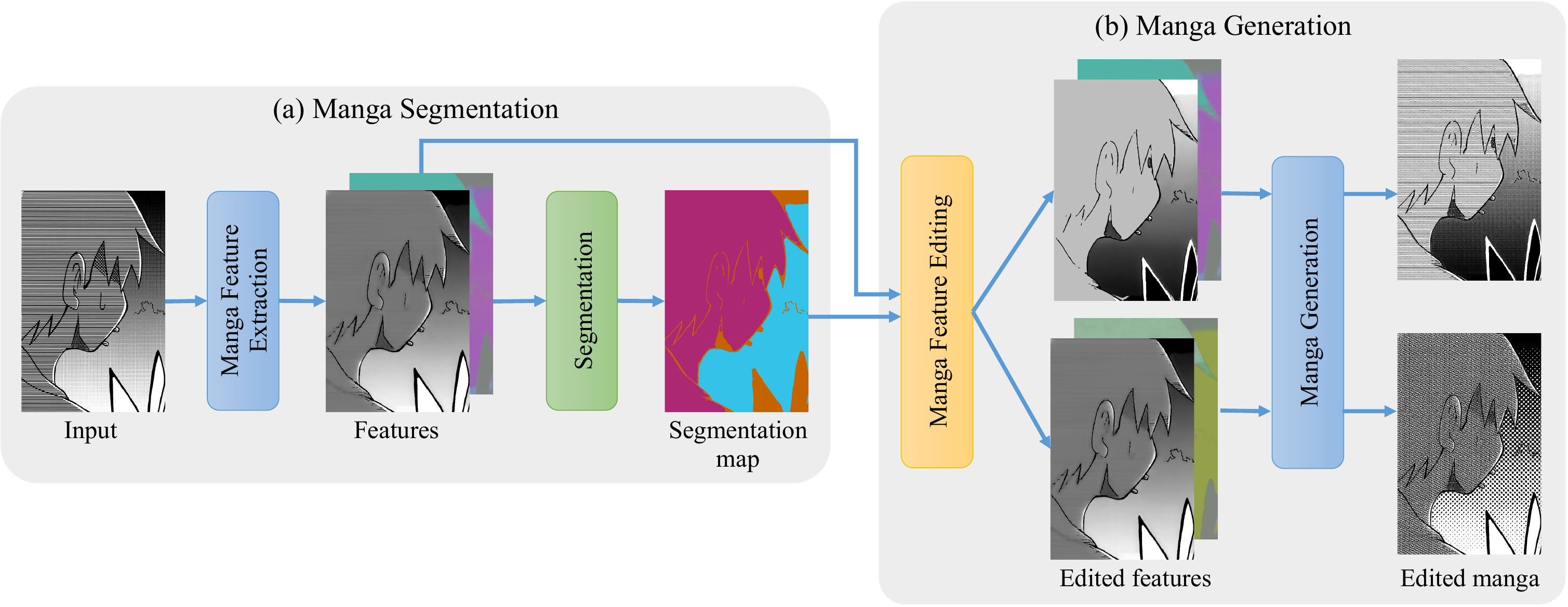}
    \caption{\textbf{Manga rescreening pipeline.} Given a manga image, we first extract the manga features and get the segmentation map with different screentone regions. We can then edit manga with user-expected characteristics, including intensity (upper branch in (b)) or screentone type (lower branch in (b)). }
    \vspace{-.1in}
    \label{fig:pipeline}
\end{figure*}

\paragraph{Representation Learning and Disentanglement}
Variational AutoEncoders (VAE) \cite{kingma2013auto} and Generative Adversarial Networks (GAN) \cite{goodfellow2014generative} are two of the most popular frameworks for representation learning. VAE learns a bidirectional mapping between complex data distribution and a much simpler prior distribution while GAN learns a unidirectional mapping that only allows sampling of data distribution. 
Disentanglement is a useful property in representation learning which increases the interpretability of the latent space, connecting certain parts of the latent representation to semantic factors, which would enable a more controllable and interactive generation process. InfoGAN \cite{chen2016infogan} disentangles latent representation by encouraging the mutual information between the observation and a small subset of the latent variables. $\beta$-VAE \cite{higgins2016beta} and some follow-up studies \cite{chen2018isolating,kim2018disentangling,yang2019inspecting} introduced various extra constraints and properties on the prior distribution. However, the above disentanglement is implicit. Though the model separates the latent space into subparts, we cannot define their meanings beforehand. 
On the contrary, some approaches aim at controllable image generation with explicit disentanglement \cite{yingzhen2018disentangled,karras2019style}. Disentangled Sequential Autoencoder \cite{yingzhen2018disentangled} learns the latent representation of high dimensional sequential data, such as video or audio, by splitting it into static and dynamic parts with a partially time-invariant encoder. 
StyleGAN \cite{karras2019style} defined the meanings of different parts of the latent representation by the model structure, making the generation controllable and more precise. 
In this paper, we attempt to disentangle the type and intensity information in the latent representation, so that the screentone generation can be controllable with type and intensity.

\section{Overview}

As highlighted in Sec.~\ref{sec:intro}, the key to intuitive user-friendly rescreening is to train an interpretable screentone representation that explicitly encodes the complex screentone intensity and type features into a low-frequency latent space. The orthogonal encoding of screentone type and intensity empowers the model to recognize screentone type similarity or equivalence without being confused by the variation of screentone intensity, and vice versa. More importantly, the representation is invertible, so that one can easily modify the low-frequency representation to generate new screentones with desired screen type or intensity. 
To enable this controllable and user-friendly encoding, we build the whole framework upon a disentangled VAE model. Different from ScreenVAE~\cite{xie2020manga}, we explicitly enforce one of the feature descriptors to encode the intensity of manga, and the screentones with the same intensity are mapped to space following a normal distribution. 
We will describe our interpretable representation in Sec.~\ref{sec:app} in more detail.

\begin{figure*}[t!]
    \centering
    \includegraphics[width=.95\linewidth]{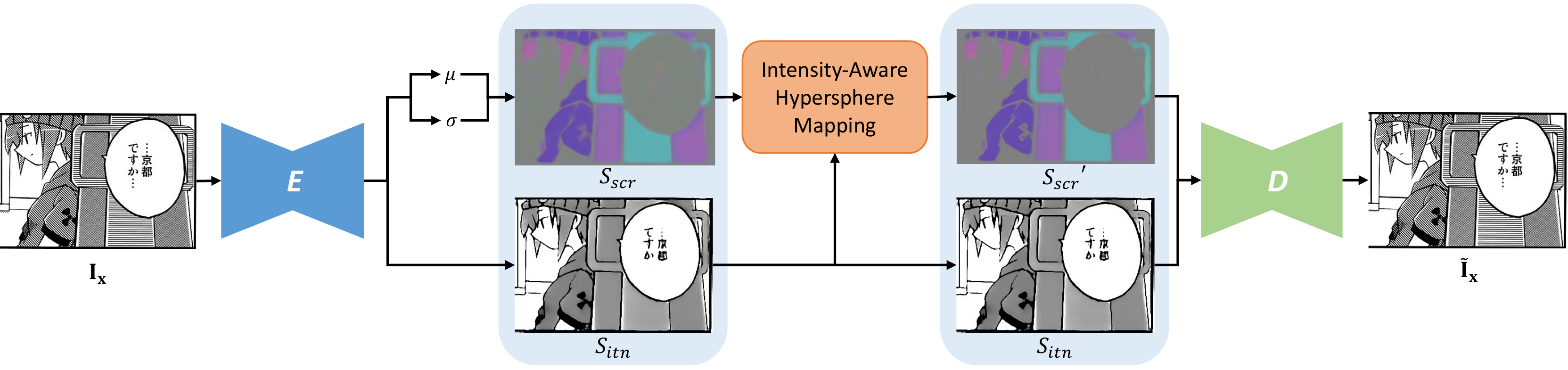}
    \caption{\textbf{Overview diagram of our screentone representation learning.} The encoder $E$ converts any screened manga $X$ to a latent representation $S$, consisting of intensity feature $S_{\rm itn}$ and screen type feature $S_{\rm scr}$. The decoder $D$ converts any latent representation $S$ to their original screened appearance $\tilde{X}$. }
    \vspace{-.1in}
    \label{fig:overview}
\end{figure*}

With the screentone representation learned, we propose our manga rescreening framework in two stages (detailed in Sec.~\ref{sec:pipeline}). In the first stage, identifies semantic continuous regions based on the screen type similarity and segments them with a Gaussian mixture model (GMM) analysis (Fig.~\ref{fig:pipeline}(a)). With the region segmented, for each screentone region, the users can directly edit the latent representation to alter either the screentone intensity (Fig.~\ref{fig:pipeline}(b) upper branch) or the screentone type (Fig.~\ref{fig:pipeline}(b) lower branch).

\section{Interpretable Screentone Representation}
\label{sec:app}
Fig.~\ref{fig:overview} illustrates our framework for learning interpretable screentone representation, which consists of two jointly trained networks, an encoding neural network $E$ and a decoding neural network $D$. The encoder $E$ converts any screened manga $X\in \mathbb{R}^{1\times H \times W}$ to a latent representation $S\in \mathbb{R}^{K\times H \times W}$, where $K$ is the dimensionality of each latent embedding vector. On the contrary, the decoder $D$ converts any latent representation to its original screened appearance $\tilde{X}$. The latent representation is defined as 4 dimensions ($K=4$) including one dimension for the intensity of screentone $S_{\rm itn}\in \mathbb{R}^{1\times H \times W}$ and three for the type of screentone $S_{\rm scr}\in \mathbb{R}^{3\times H \times W}$. 
Besides, variational inference~\cite{kingma2013auto} is employed to ensure the latent representation to be interpolative. The encoder $E$ adopts a 3-level downscaling-upscaling structure with 6 residual blocks \cite{he2016deep} and the decoder $D$ adopts a 7-level U-net structure \cite{ronneberger2015u} with strided deconvolution operations to generate screentones of different scales.

To improve the generalization and fully disentangle the intensity, we impose an extra path by introducing a random intensity map. A random latent representation is generated by combining the encoded latent type feature $S_{\rm scr}$ and random intensity map $S_{\rm ritn}$. Given the random latent representation $S_r$, the decoder $D$ is expected to generate a realistic image $\tilde{X_r}$, and the reconstructed latent representation $\tilde{S_r}$ generated by the encoder $E$ should also be as similar as the given latent representation $S_r$.

\begin{figure}[t!]
    \centering
    \includegraphics[width=\linewidth]{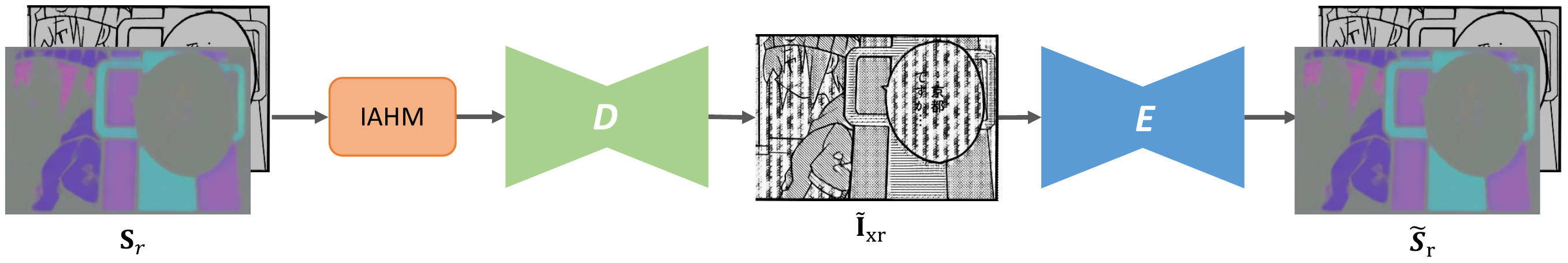}
    \caption{Given the random latent representation $S_r$, the decoder can generate a realistic image $\tilde{X_r}$ which will then reconstruct a similar latent representation $\tilde{S_r}$ by the encoder. }
    \vspace{-.1in}
    \label{fig:randomintensity}
\end{figure}

\subsection{Intensity-Aware Hypersphere Modeling}

\begin{figure}[t!]
    \centering
    \includegraphics[width=.8\linewidth]{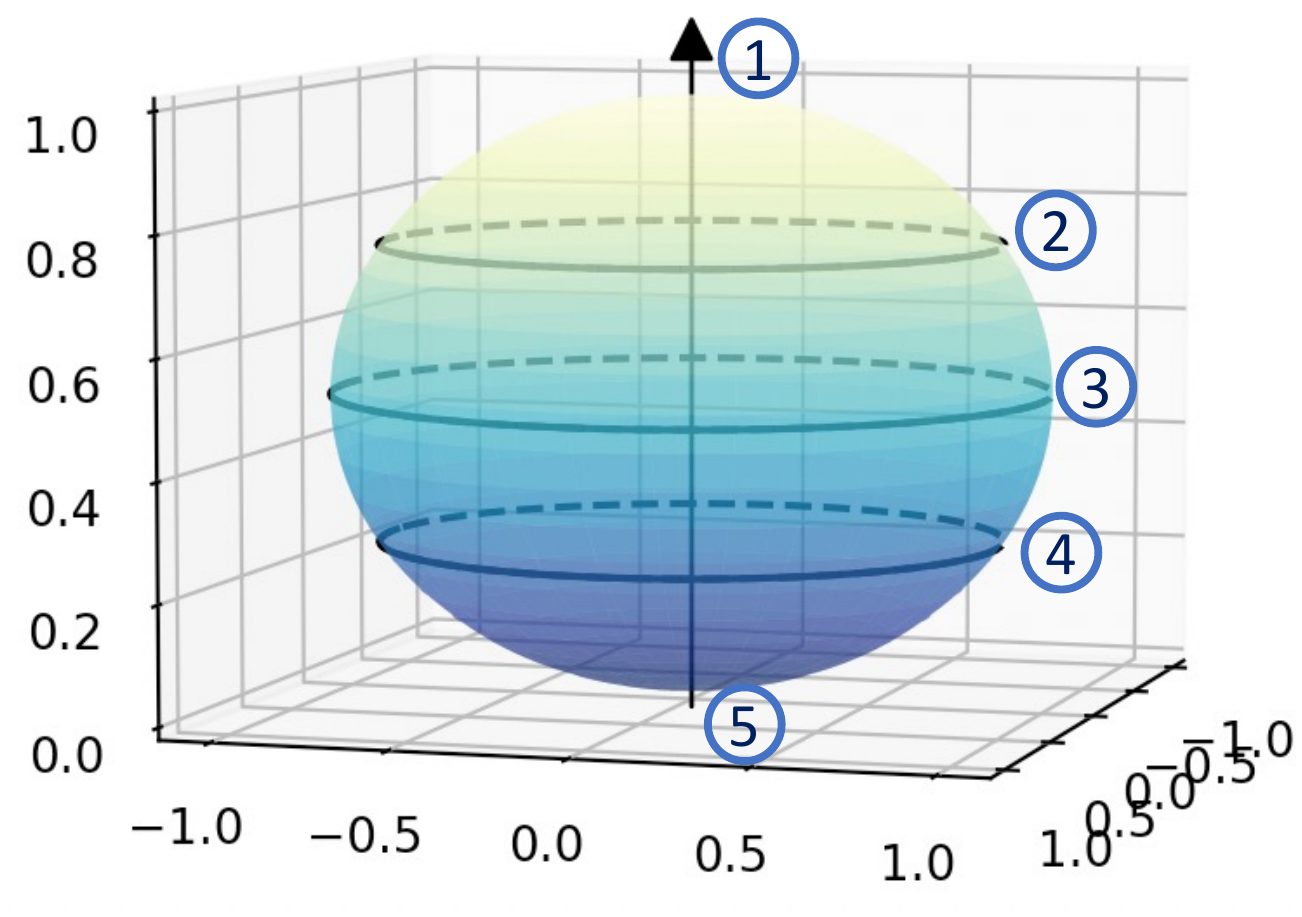}\\
    \makebox[0.4\linewidth]{(a) Latent space with intensity as one axis}\hfil
    \includegraphics[width=.95\linewidth]{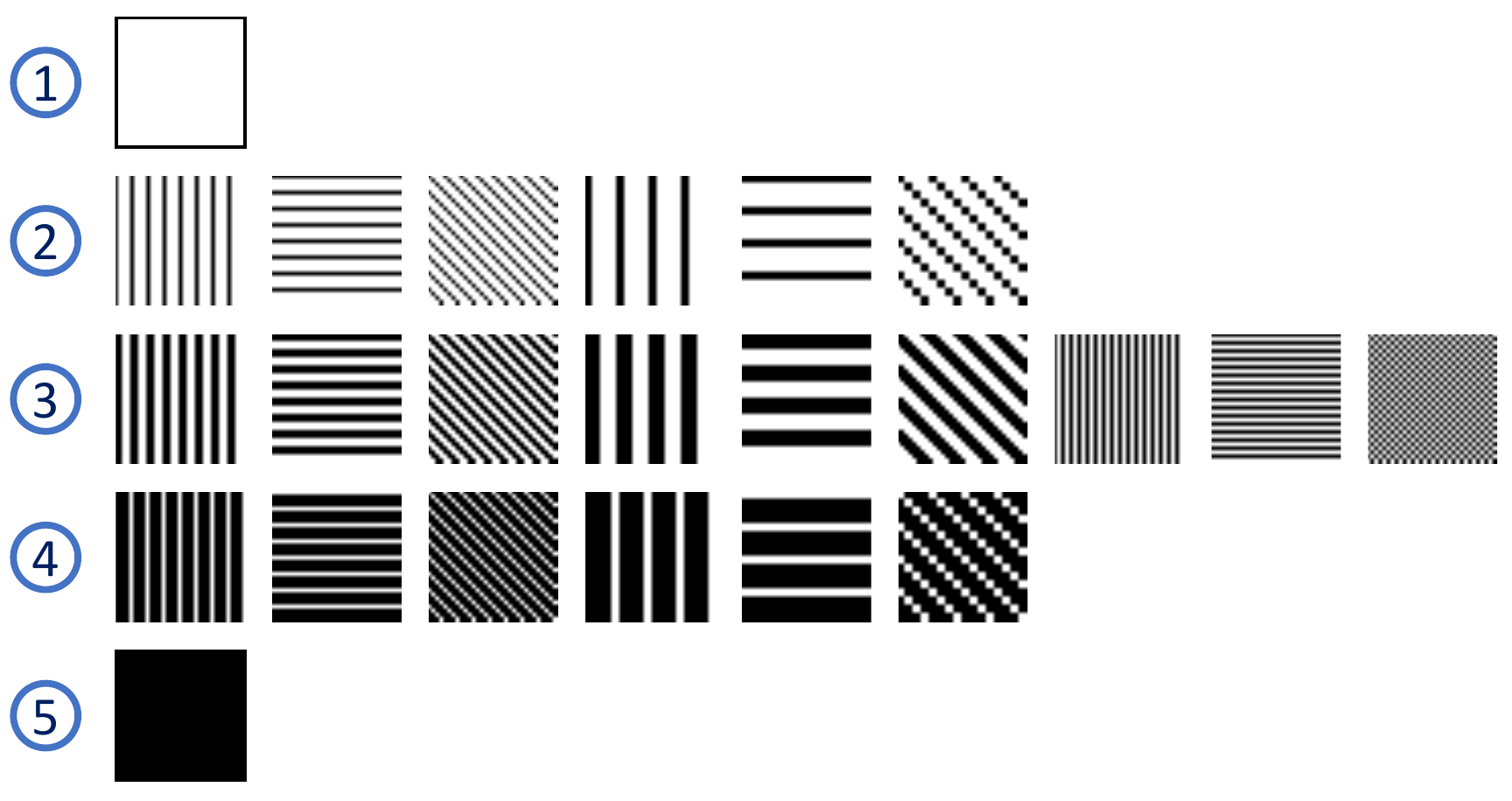}\\
    \makebox[0.55\linewidth]{(b) Example screentones with different intensity}\hfil
    \caption{The diversity of screentones is correlated with the intensity, which decreases from 50\% intensity to pure white or black intensity. Note that z axis represents intensity while x-y axis features the types and is from a PCA-ed space. }
    \label{fig:tonespace}
\end{figure}

We observed that the diversity of screentones with different intensities has the following properties. Firstly, the domain conditioned on intensity should be a symmetric space, as each pattern can transform into a new pattern with opposite intensity by inverting the black and white pixels. Second, the screentone diversity is correlated with its intensity. The diversity gradually decreases when annihilated into a pure black or pure white pattern from 50\% intensity, as shown in Fig.~\ref{fig:tonespace}(b). In particular, when the intensity is 100\% or 0\%, there is no variation of screentones. Considering the above properties, we propose to model the domain as a hypersphere with intensity as one \textbf{independent} axis, as illustrated in Fig.~\ref{fig:tonespace}(a). 

Intensity-Aware Hypersphere Mapping (IAHM) is proposed to achieve the modeling. Considering the diversity of the screen types is conditioned by the intensity information, we hereby model the types of screentones with the same intensity as a normal distribution $\mathcal{N}(\textbf{0},r^2\cdot \textbf{I})$ instead of projecting all screentones to a standard normal distribution in high dimensions. Meanwhile, we force the embedding domain to be hyperspherical with constraints of $r=f(S_{\rm itn})\sim\sin(\pi\times S_{\rm itn})$. As we can have $\frac{1}{r}\cdot S_{\rm scr}' \sim\mathcal{N}(\textbf{0, I})$ with $S_{\rm scr}' \sim\mathcal{N}(\textbf{0},r^2\cdot \textbf{I})$, our Intensity-Aware Hypersphere Mapping (IAHM) is then defined as  
\begin{equation}
    S_{\rm scr}' = r\odot S_{\rm scr}, {\rm where}\  r=\sin(\pi\times S_{\rm itn})
\end{equation}
where $\odot$ indicates point-wise multiplication. In particular, it is deterministic with $r=0$ at black and white intensity. 

\begin{figure}
    \centering
    \settoheight{\tempdim}{\includegraphics[width=1.9\linewidth]{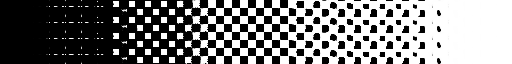}}%
    \rotatebox{90}{\makebox[\tempdim]{\small (a) w/o IAHM}}\hfil
    \begin{minipage}[b]{.95\linewidth}
        \includegraphics[width=\linewidth]{imgs/ablation/iahm/wo_iahm_512.png}
        \begin{minipage}[b]{\linewidth}
            \includegraphics[width=.11\linewidth]{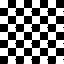}
            \includegraphics[width=.11\linewidth]{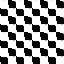}
            \includegraphics[width=.11\linewidth]{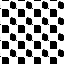}
            \includegraphics[width=.11\linewidth]{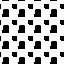}
            \includegraphics[width=.11\linewidth]{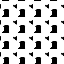}
            \includegraphics[width=.11\linewidth]{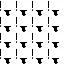}
            \includegraphics[width=.11\linewidth]{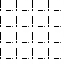}
            \includegraphics[width=.11\linewidth]{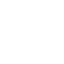}
        \end{minipage}
    \end{minipage}
    \rotatebox{90}{\makebox[\tempdim]{\small (b) w IAHM}}\hfil
    \begin{minipage}[b]{.95\linewidth}
        \includegraphics[width=\linewidth]{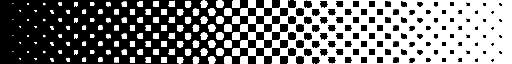}
        \begin{minipage}[b]{\linewidth}
            \includegraphics[width=.11\linewidth]{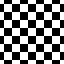}
            \includegraphics[width=.11\linewidth]{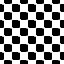}
            \includegraphics[width=.11\linewidth]{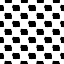}
            \includegraphics[width=.11\linewidth]{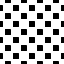}
            \includegraphics[width=.11\linewidth]{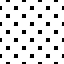}
            \includegraphics[width=.11\linewidth]{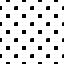}
            \includegraphics[width=.11\linewidth]{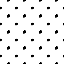}
            \includegraphics[width=.11\linewidth]{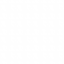}
        \end{minipage}
    \end{minipage}
    \caption{Our method with IAHM can generate smooth transitions with the same type of screentones. }
    \label{fig:withoutIAHM}
    \vspace{-0.1in}
\end{figure}

The IAHM encoding substantially improves the usage efficiency of latent space, which helps to avoid the model bias due to imbalanced varieties of different intensities. As shown in Fig.~\ref{fig:withoutIAHM}, the model without IAHM will not be able to generate screentones with darker or lighter tone intensity. Specifically, for some screentones with darker or lighter intensity, although they may be quite similar at the pixel level, they may have a large distance in the latent space without the proposed IAHM, which will make the model fail to learn these screentones. 

\subsection{Loss Function}
Our model is trained with the loss function defined in Equ.\ref{equ:loss}, consisting of reconstruction loss $\mathcal{L}_{\rm rec}$, adversarial loss $\mathcal{L}_{\rm adv}$, intensity loss $\mathcal{L}_{\rm itn}$, KL divergence loss $\mathcal{L}_{\rm kl}$, feature consistency loss $\mathcal{L}_{\rm fcons}$, and feature reconstruction loss $\mathcal{L}_{\rm frec}$. The objective is formulated as
\begin{equation}
    \begin{array}{ll}
     \mathcal{L} = & \lambda_{\rm rec}\mathcal{L}_{\rm rec} + \lambda_{\rm adv}\mathcal{L}_{\rm adv} + \lambda_{\rm itn}\mathcal{L}_{\rm itn} \\
     & + \lambda_{\rm kl}\mathcal{L}_{\rm kl} + \lambda_{\rm fcons}\mathcal{L}_{\rm fcons} + \lambda_{\rm frec}\mathcal{L}_{\rm frec}
    \end{array}
    \label{equ:loss}
\end{equation}
where the hyper-parameters $\lambda_{\rm rec}$ = 10, $\lambda_{\rm adv}$ = 1, $\lambda_{\rm itn}$ = 5, $\lambda_{\rm kl}$ = 1, $\lambda_{\rm fcons}$ = 20, and $\lambda_{\rm frec}$ = 1 are set empirically.

\textbf{Reconstruction loss. }
The reconstruction loss $\mathcal{L}_{\rm rec}$ ensures that the reconstructed manga $\tilde{X}$ is as similar to the input $X$ as possible, formulated in pixel level. The reconstruction loss is defined as 
\begin{equation}
    \mathcal{L}_{\rm rec}=\|\tilde{X}-X\|_2^2
\end{equation}
where $\|\cdot\|_2$ denotes the $L_2$ norm.

\textbf{Adversarial loss. }
The adversarial loss $\mathcal{L}_{\rm adv}$ encourages the decoder to generate manga with clear and visually pleasant screentones. We treat our model as a generator and define an extra discriminator $D_{\rm m}$ with 4 strided downscaling blocks to judge whether the input image is generated or not~\cite{goodfellow2014generative}. Given the input image $X$ and the reconstructed image $\tilde{X}$ and $\tilde{X_r}$, we formulate the adversarial loss as 
\begin{equation}
    \mathcal{L}_{\rm adv}=\log(D_{\rm m}(X)) + \log(1-D_{\rm m}(\tilde{X}))+ \log(1-D_{\rm m}(\tilde{X_r})).
\end{equation}

\textbf{Intensity loss. }
The intensity loss $\mathcal{L}_{\rm itn}$ ensures the generated intensity map $S_{\rm itn}$ visually conforms to the intensity $I_{\rm itn}$ of the manga image. We formulate it as
\begin{equation}
    \mathcal{L}_{\rm itn} = \|S_{\rm itn}-I_{\rm itn}\|_2^2.
\end{equation}

\textbf{KL divergence loss. }
The KL divergence loss $\mathcal{L}_{\rm kl}$ ensures that the statistics of the type feature $S_{\rm scr}$ are normally distributed. Given an input manga $X$ and the encoded representation $\mu$ and $\sigma$, we compute the KL divergence loss as the summed Kullback-Leibler divergence~\cite{kullback1951information} of $\mu$ and $\sigma$ over the standard normal distribution $\mathcal{N}(\mathbf{0,I})$. 
\begin{equation}
    \begin{array}{ll}
        \mathcal{L}_{\rm kl}&=KL(\mathcal{N}(\mu,\sigma),\mathcal{N}(\mathbf{0,I}))\\
        &=\frac{1}{2}\sum(\sigma^2+\mu^2-\log(\sigma^2)-1)
    \end{array}
\end{equation}
Here, $KL(\cdot,\cdot)$ denotes the KL divergence between two probability distributions.

\textbf{Feature consistency loss. }
The feature consistency loss $\mathcal{L}_{\rm fcons}$ ensures that the type feature $S_{\rm scr}$ can summarize the local texture characteristics in the input $X$~\cite{xie2020manga}. 
Given the type feature $S_{\rm scr}$ and its label map $I_{\rm lb}$ which labels the screentone of each pixel, we encourage a uniform region representation within the region filled with the same screentone~\cite{kwak2017weakly}. The feature consistency loss is formulated as 
\begin{equation}
    \mathcal{L}_{\rm fcons}=\|w_l \cdot (S_{\rm scr}-{\rm Avg}(S_{\rm scr},I_{\rm lb}))\|_2^2
\end{equation}
where $w_l$ is the binary mask to filter out structure lines (0 for structural lines). ${\rm Avg}(S_{\rm scr}, I_{\rm lb})$ is a map in which each pixel is replaced by the average value of the corresponding region indexed by $I_{\rm lb}$ in the representation $S_{\rm scr}$. 

\textbf{Feature reconstruction loss. }
We find that intensity information may still entangle with type feature when only the above losses are imposed. 
To disentangle intensity feature from type feature, we encourage the type feature can still be reconstructed through the extra path with random intensity map. The feature reconstruction loss $\mathcal{L}_{\rm frec}$ is measured as the pixel-wise difference between the random representation $S_r$ and the reconstructed representation $\tilde{S_r}$. 
\begin{equation}
    \mathcal{L}_{\rm frec} = \|\tilde{S_r}-S_r\|_2^2
\end{equation}

\section{Experiments}

\begin{figure*}[th!]
    \centering
    \subfloat[Original manga]{
    \begin{minipage}[c]{.16\linewidth}
        \adjincludegraphics[clip,trim=0 {.15\width} 0 {.15\width},width=\linewidth]{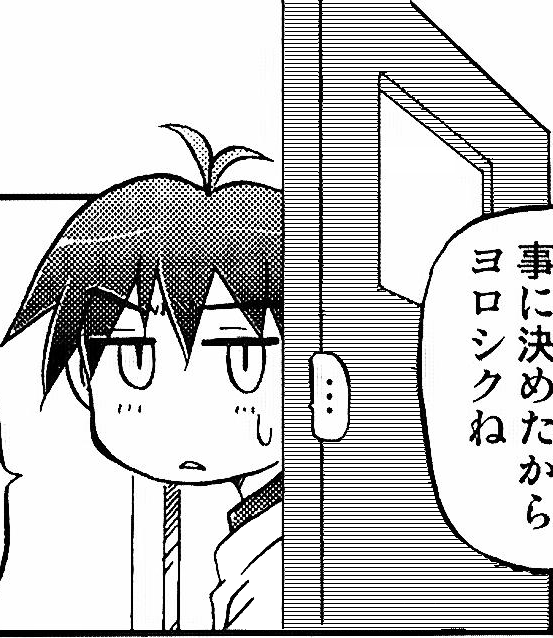}
        \includegraphics[width=\linewidth]{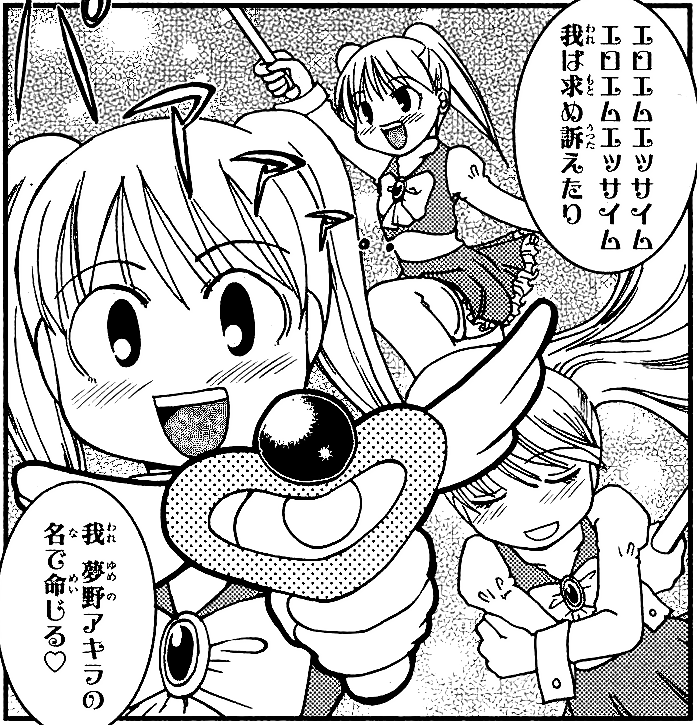}
        \includegraphics[width=\linewidth]{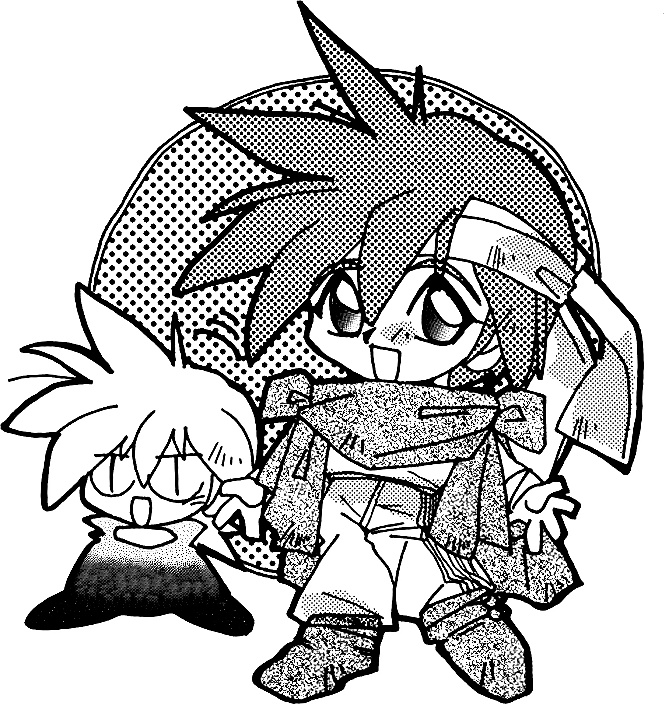}
    \end{minipage}}
    \subfloat[Intensity feature]{
    \begin{minipage}[c]{.16\linewidth}
        \adjincludegraphics[clip,trim=0 {.15\width} 0 {.15\width},width=\linewidth]{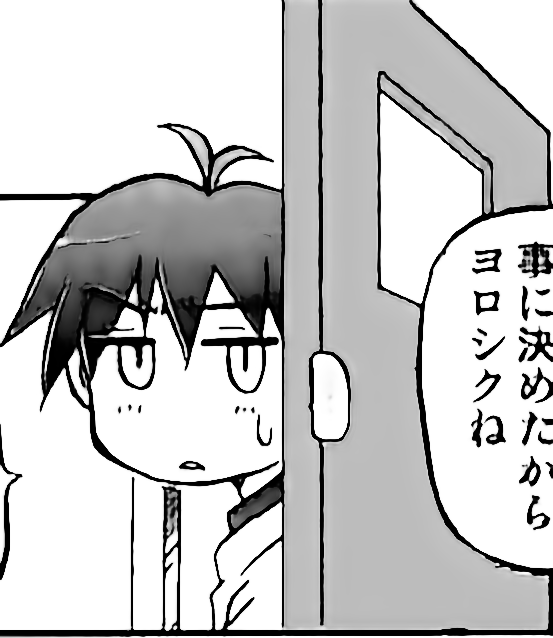}
        \includegraphics[width=\linewidth]{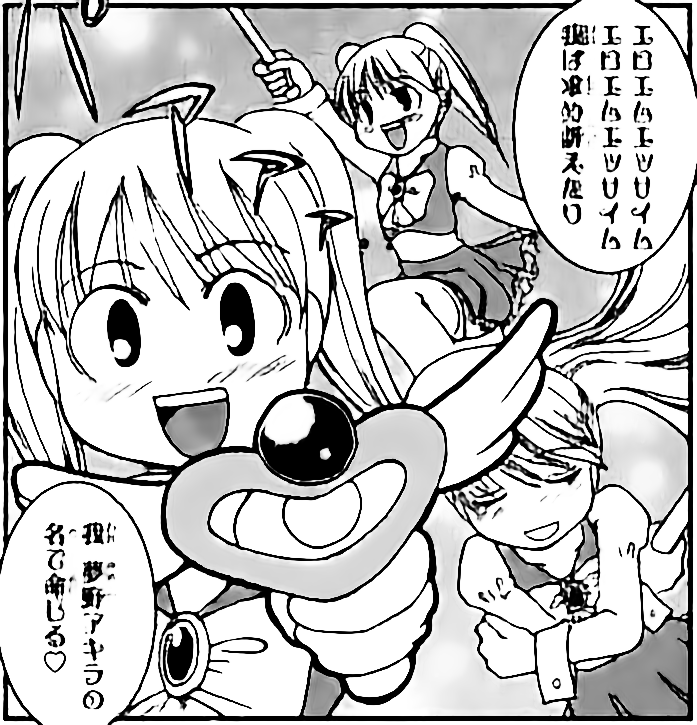}
        \includegraphics[width=\linewidth]{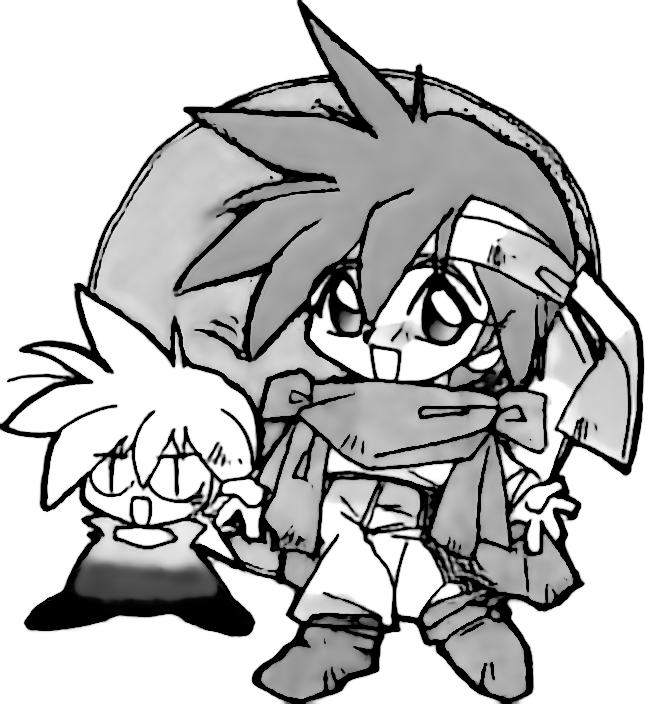}
    \end{minipage}}
    \subfloat[Type feature]{
    \begin{minipage}[c]{.16\linewidth}
        \adjincludegraphics[clip,trim=0 {.15\width} 0 {.15\width},width=\linewidth]{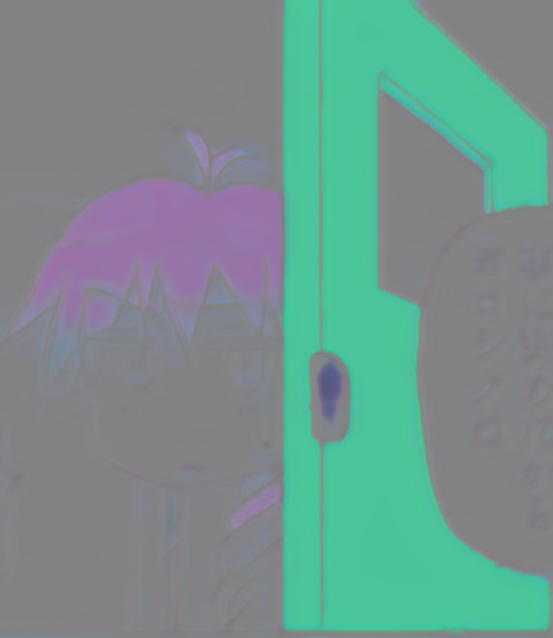}
        \includegraphics[width=\linewidth]{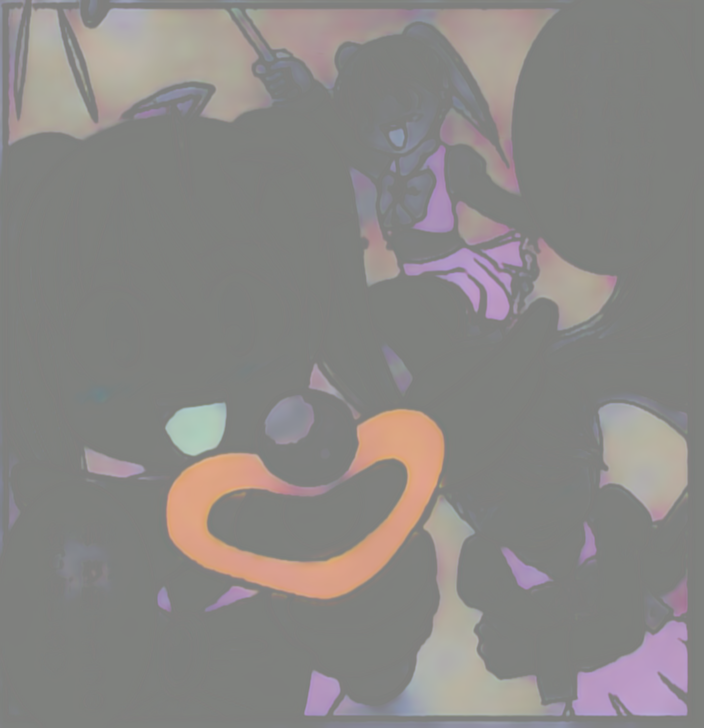}
        \includegraphics[width=\linewidth]{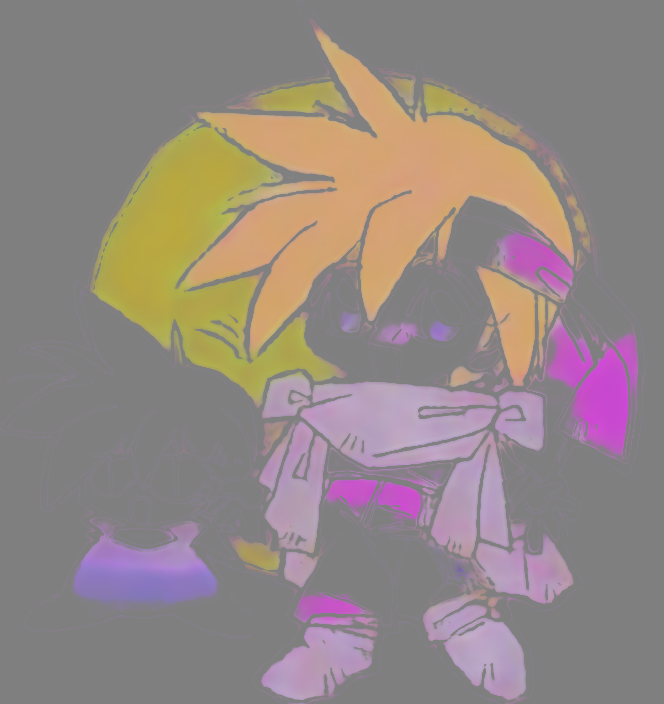}
    \end{minipage}}
    \subfloat[Segmentation map]{
    \begin{minipage}[c]{.16\linewidth}
        \adjincludegraphics[clip,trim=0 {.15\width} 0 {.15\width},width=\linewidth]{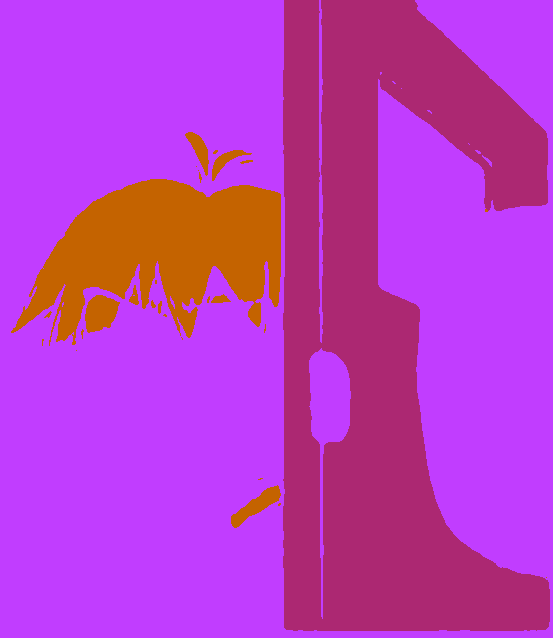}
        \includegraphics[width=\linewidth]{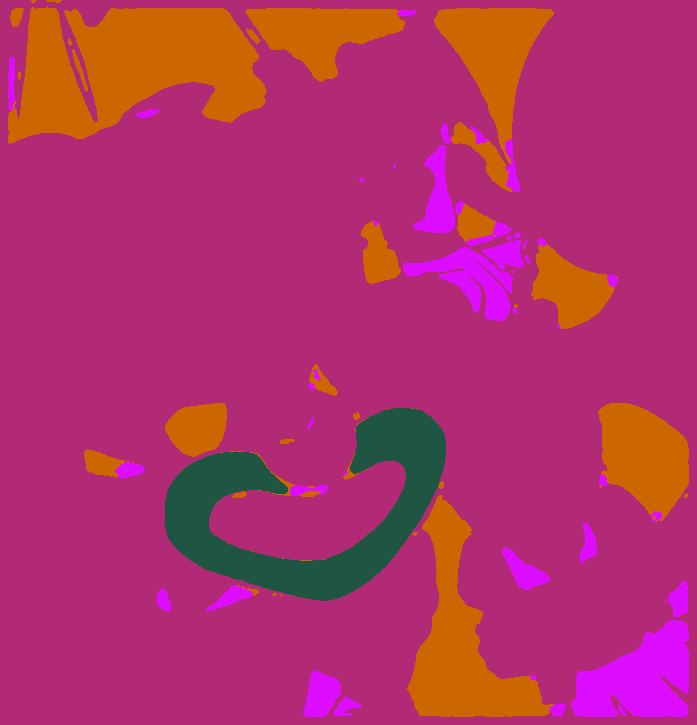}
        \includegraphics[width=\linewidth]{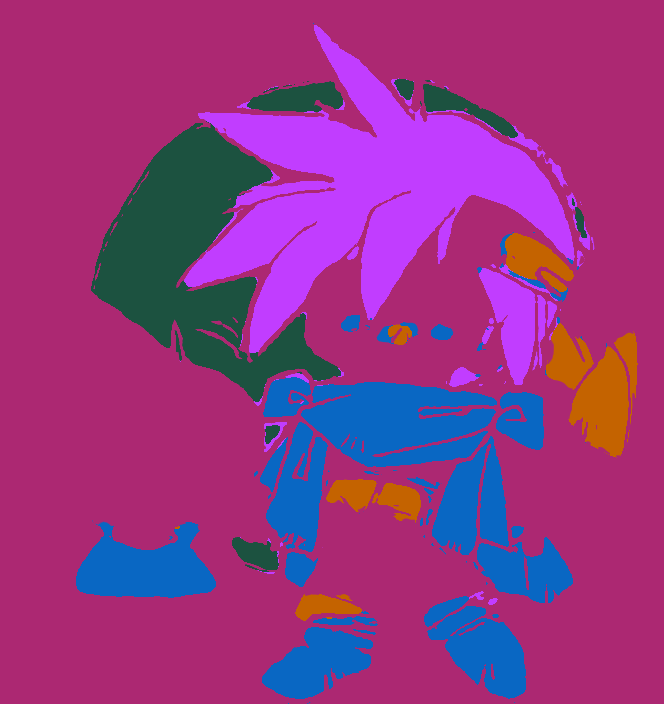}
    \end{minipage}}
    \subfloat[Edited with type]{
    \begin{minipage}[c]{.16\linewidth}
        \adjincludegraphics[clip,trim=0 {.15\width} 0 {.15\width},width=\linewidth]{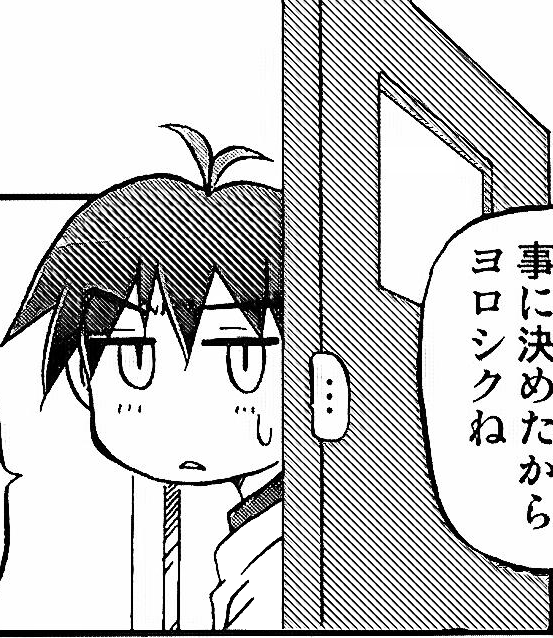}
        \includegraphics[width=\linewidth]{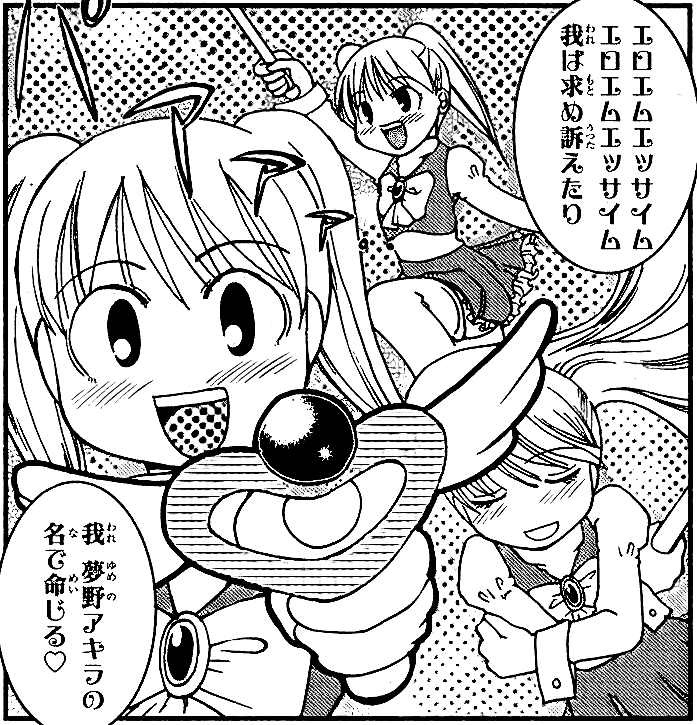}
        \includegraphics[width=\linewidth]{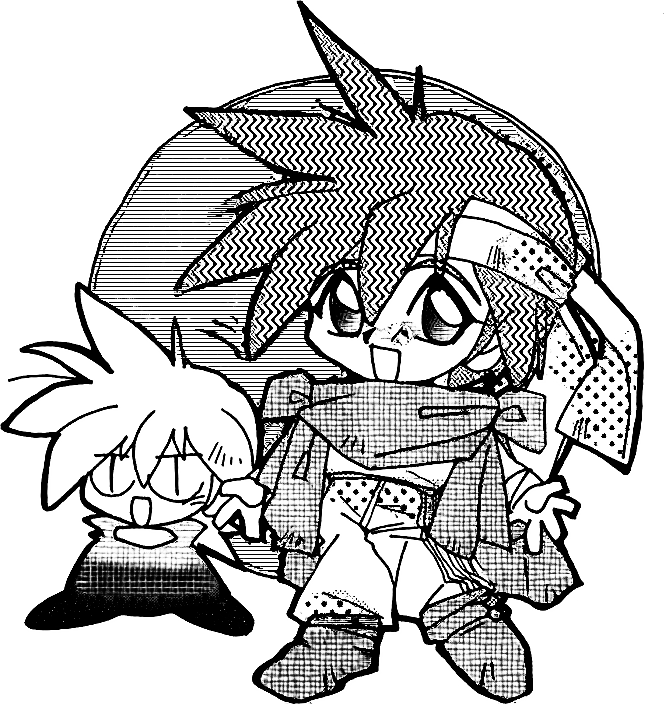}
    \end{minipage}}
    \subfloat[Edited with intensity]{
    \begin{minipage}[c]{.16\linewidth}
        \adjincludegraphics[clip,trim=0 {.15\width} 0 {.15\width},width=\linewidth]{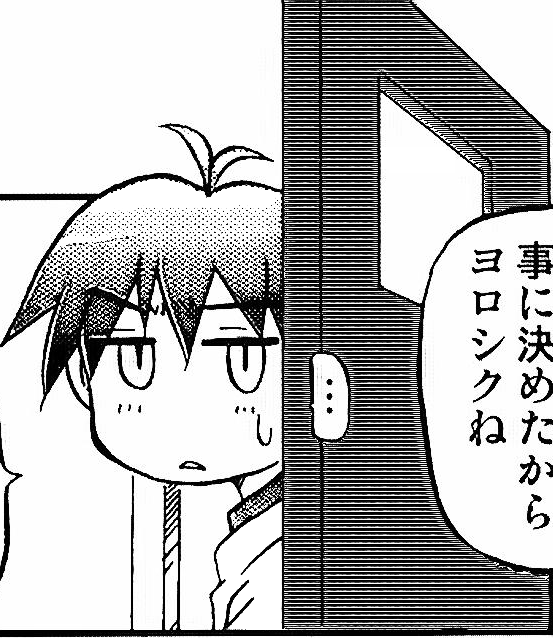}
        \includegraphics[width=\linewidth]{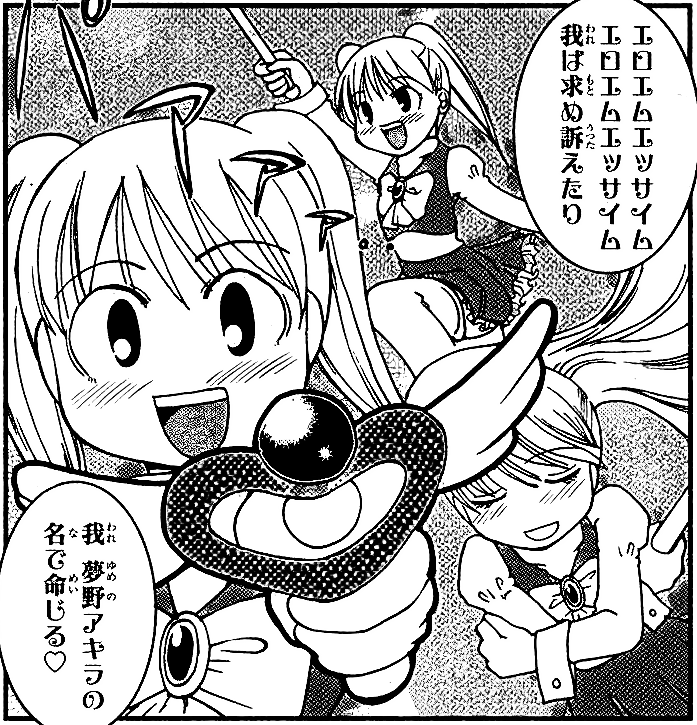}
        \includegraphics[width=\linewidth]{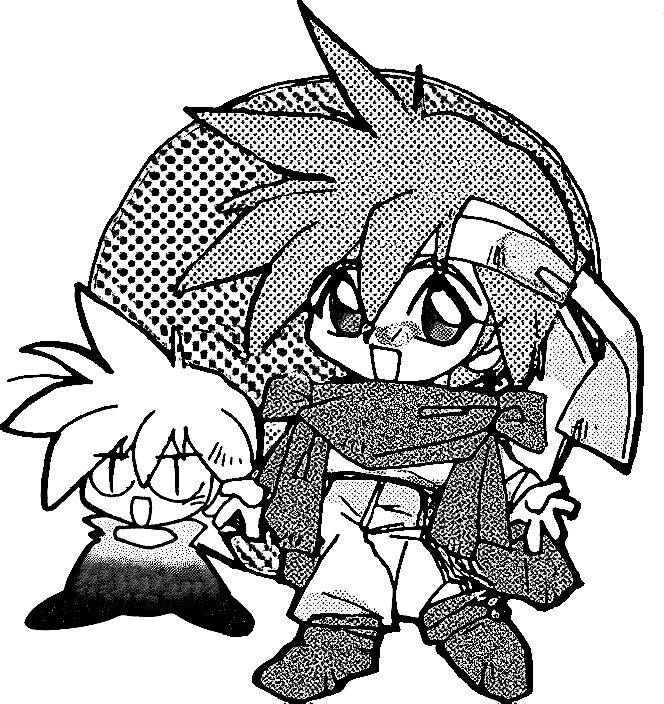}
    \end{minipage}}
    \vspace{-.05in}
    \caption{Our method can extract screentone regions and cover them with new screentones. Akuhamu \textcopyright Arai Satoshi, ReveryEarth \copyright Yama Miyuki~\cite{mtap_matsui_2017}}
    \vspace{-.15in}
    \label{fig:application}
\end{figure*}

\subsection{Implementation Details}
\textbf{Data Preparation. }
We use two types of data to train our model, including synthetic manga data and real manga data. 
For the synthetic manga data, we manually collected 100 types of screentones and generated their tone inverse by swapping the black and white pixels. We then collected 500 line arts and generated synthetic manga following Li et al.~\cite{li2017deep} which randomly choose and place screentones in each closed region. We synthesized 5,000 manga images of resolution 2,048$\times$ 1,536, together with the intensity and the screentone type labels of each pixel. Intensity maps and label maps are used to calculate intensity loss $\mathcal{L}_{\rm itn}$ and feature consistency loss $\mathcal{L}_{\rm fcons}$, respectively. For the real manga data, we manually collected 5,000 screened manga of resolution 2,048$\times$ 1,536 to train our model. For each screened manga, we extract the structural lines using the line extraction model~\cite{li2017deep} and the intensity maps using total-variation-based smoothing~\cite{xu2011image}. Note that we do not label their ground truth screen type, so the feature consistency loss $\mathcal{L}_{\rm fcons}$ is not calculated for this portion of data. 

\textbf{Training. }
We trained the model with PyTorch~\cite{paszke2017automatic}. The network weights are initialized with~\cite{he2015delving}. During training, we empirically set parameters as $\lambda_{\rm rec}$ = 10, $\lambda_{\rm adv}$ = 1, $\lambda_{\rm itn}$ = 5, $\lambda_{\rm kl}$ = 1, $\lambda_{\rm fcons}$ = 20 and $\lambda_{\rm frec}$ = 1. Adam solver~\cite{kingma2014adam} is used with a batch size of 8 and an initial learning rate of 0.0001. All image pairs are cropped to $256\times 256$ before feeding to the networks.
Considering the bias problem of real data, the whole model is first trained with synthetic data to obtain a stable latent space. Then, both synthetic and real data are imposed for training to improve generalization. 

\subsection{The Rescreening Pipeline}
\label{sec:pipeline}
With the learned representation of screentones, users can easily edit the screentones in manga by segmenting the screentone region and generating new screentones. 

\begin{figure*}[th!]
    \centering
    \subfloat[Input manga]{
    \begin{minipage}[c]{.14\textwidth}
        \includegraphics[width=\linewidth]{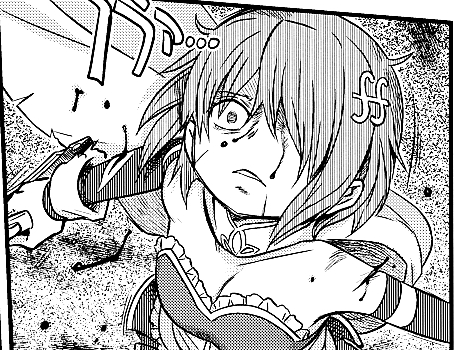}
        \adjincludegraphics[clip,trim=0 {.2\height} {.1\width} {.1\height},width=\linewidth]{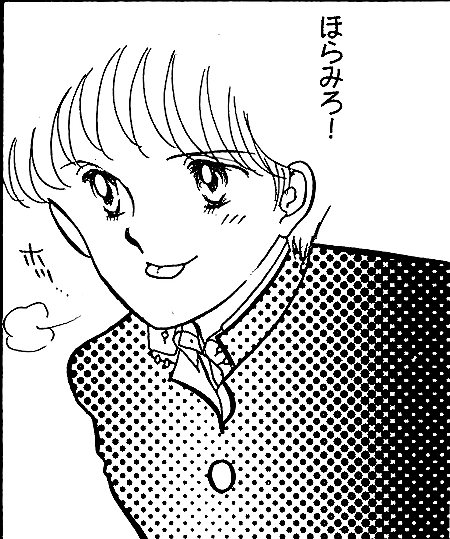}
    \end{minipage}}
    \subfloat[\cite{manjunath1996texture}]{
    \begin{minipage}[c]{.14\textwidth}
        \includegraphics[width=\linewidth]{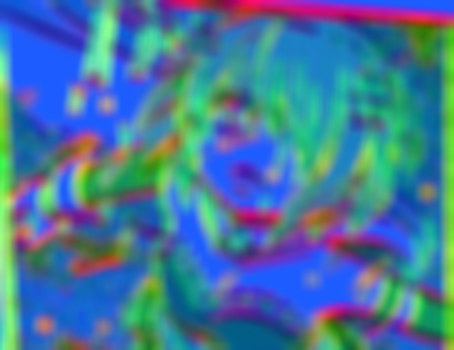}
        \adjincludegraphics[clip,trim=0 {.2\height} {.1\width} {.1\height},width=\linewidth]{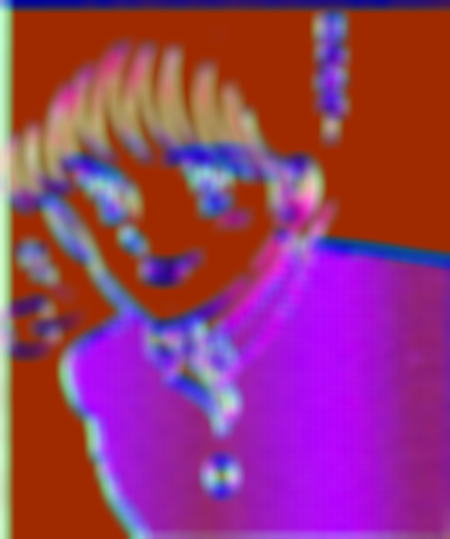}
    \end{minipage}}
    \subfloat[Segmentation of~\cite{manjunath1996texture}]{
    \begin{minipage}[c]{.14\textwidth}
        \includegraphics[width=\linewidth]{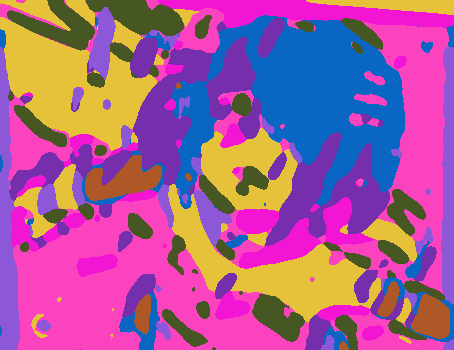}
        \adjincludegraphics[clip,trim=0 {.2\height} {.1\width} {.1\height},width=\linewidth]{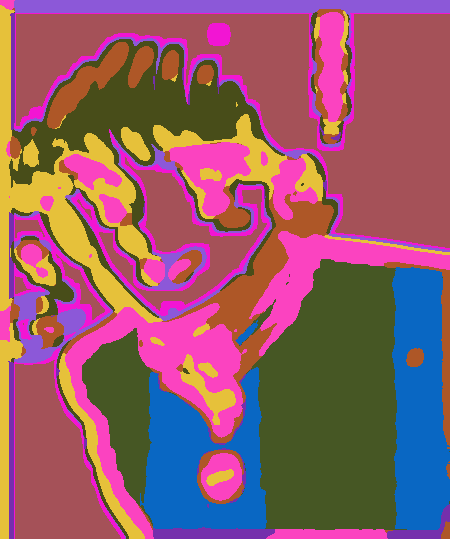}
    \end{minipage}}
    \subfloat[\cite{xie2020manga}]{
    \begin{minipage}[c]{.14\textwidth}
        \includegraphics[width=\linewidth]{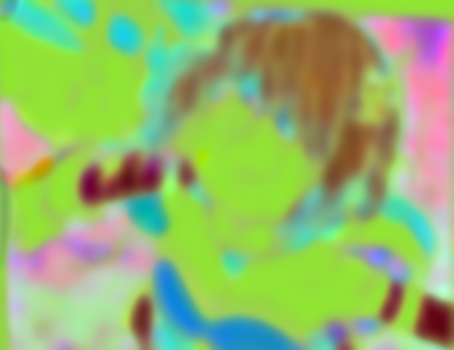}
        \adjincludegraphics[clip,trim=0 {.2\height} {.1\width} {.1\height},width=\linewidth]{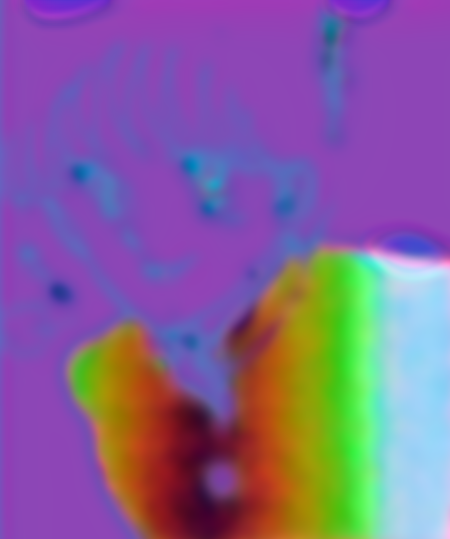}
    \end{minipage}}
    \subfloat[Segmentation of~\cite{xie2020manga}]{
    \begin{minipage}[c]{.14\textwidth}
        \includegraphics[width=\linewidth]{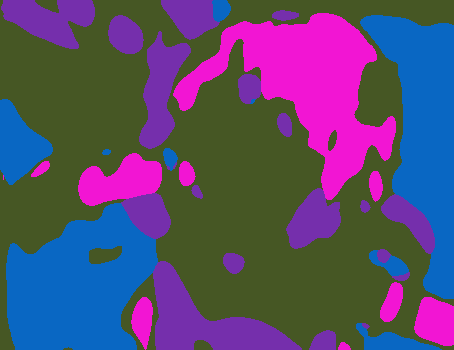}
        \adjincludegraphics[clip,trim=0 {.2\height} {.1\width} {.1\height},width=\linewidth]{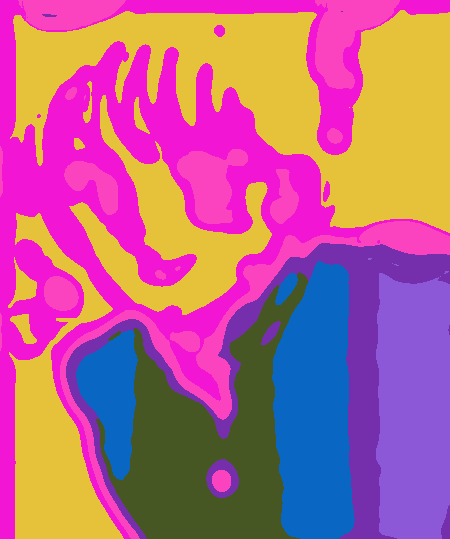}
    \end{minipage}}
    \subfloat[Our type feature]{
    \begin{minipage}[c]{.14\textwidth}
        \includegraphics[width=\linewidth]{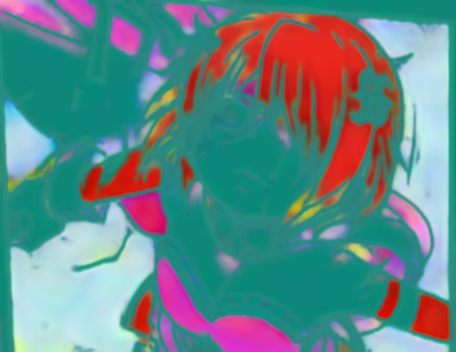}
        \adjincludegraphics[clip,trim=0 {.2\height} {.1\width} {.1\height},width=\linewidth]{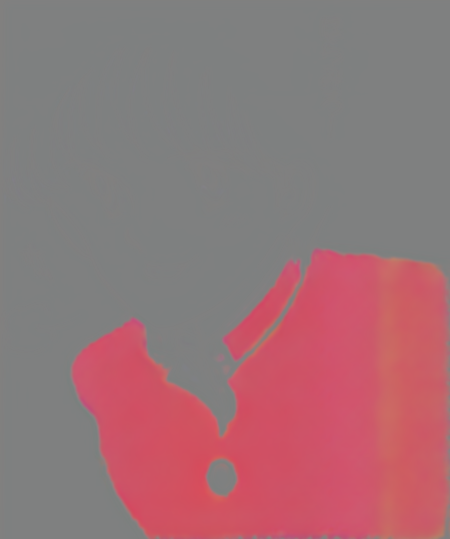}
    \end{minipage}}
    \subfloat[Our segmentation]{
    \begin{minipage}[c]{.14\textwidth}
        \includegraphics[width=\linewidth]{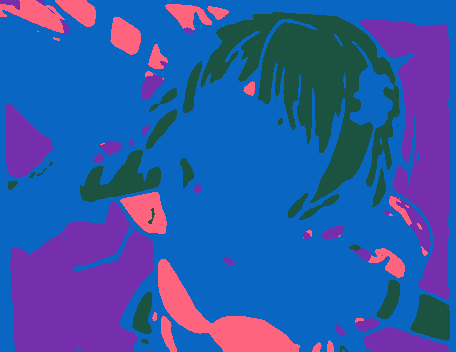}
        \adjincludegraphics[clip,trim=0 {.2\height} {.1\width} {.1\height},width=\linewidth]{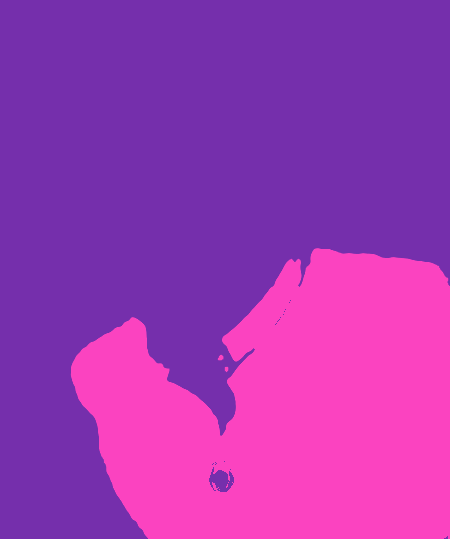}
    \end{minipage}}
    \vspace{-.07in}
    \caption{Comparison with screentone analysis methods. KaerimichiNoMajo \textcopyright A-10, SonokiDeABC \textcopyright Tashiro Kimu~\cite{mtap_matsui_2017}}
    \vspace{-.15in}
    \label{fig:comparison}
\end{figure*}

\textbf{Manga Segmentation. }
To perform automatic manga segmentation, we adopt the Gaussian mixture model (GMM) analysis~\cite{gupta1998gaussian} on the proposed representation. The optimal number of clusters is selected by the silhouette coefficient~\cite{kaoungku2018silhouette}, a clustering validation metric. As the screentone types in a single page of manga are usually limited for visual comfort, we find the optimal number of clusters ranges from 1 to 10. 
We illustrate the consistency of the encoded screen type feature for regular patterns (dot pattern in Fig.~\ref{fig:rescreening}) and noisy patterns (first row in Fig.~\ref{fig:comparison}). All features are visualized after principal component analysis (PCA)~\cite{mackiewicz1993principal}.

\textbf{Controllable Manga Generation. }
After segmentation, each segmented region can be rescreened by modifying its latent and decoding the latent back to screentones. Users can either modify the screentone type while preserving the original intensity variation (Fig.~\ref{fig:application}(e)), or the intensity feature to change the intensity (Fig.~\ref{fig:application}(f)). 
Interestingly, the editing of the intensity is very flexible. The artist can either replace an intensity-varying region with an intensity-constant one, or preserve the visual effects by increasing/decreasing the overall tone intensity proportionally, as shown in the hair region in Fig.\ref{fig:application}(f)).

\subsection{Qualitative Evaluation}
As there is no method tailored for manga rescreening, we compare with approaches on manga segmentation and controllable manga generation respectively. 

\textbf{Comparison on Manga Segmentation. }
In Fig.~\ref{fig:comparison}, we visually compare our method with the classic Gabor wavelet texture analysis~\cite{manjunath1996texture} and the learning-based ScreenVAE model~\cite{xie2020manga}. All results are visualized by considering the three major components as color values. We also apply the same segmentation to each feature by measuring texture similarity. As observed, all features have the capability of summarizing the texture characteristics in a local region and can distinguish different types of screentones. However, Gabor wavelet feature exhibits severe artifacts near region boundaries with blurry double edges due to its window-based analysis. ScreenVAE map can have tight boundaries towards structural lines, but it failed to segment regions with varying intensity (second row in Fig. \ref{fig:comparison}). In contrast, our representation explicitly encodes the intensity information and can disentangle the types. We can generate a consistent type feature for varying intensity regions.

\begin{figure}[!t]
    \centering
    \subfloat[Manga]{
    \begin{minipage}[b]{.24\linewidth}
        \includegraphics[width=.95\linewidth]{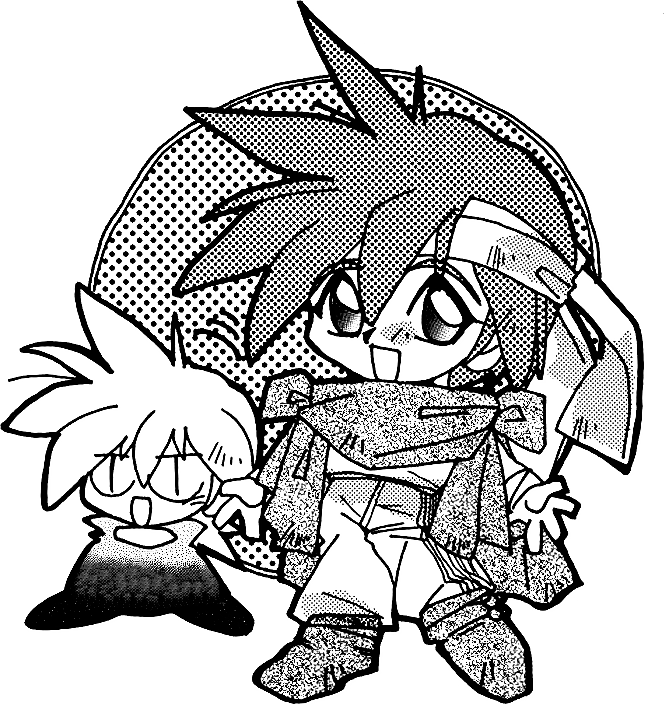}
        \adjincludegraphics[clip,trim=0 {.07\height} 0 {.15\height},width=.925\linewidth]{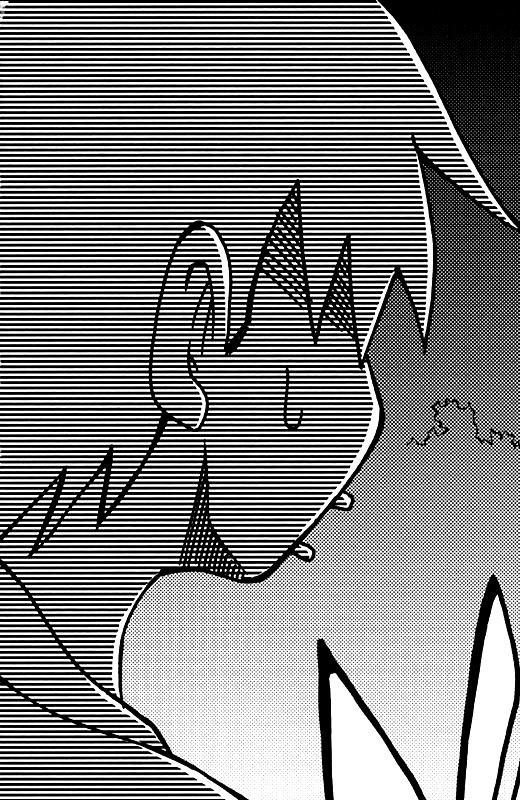}
    \end{minipage}
    }
    \subfloat[New type]{
    \begin{minipage}[b]{.24\linewidth}
        \includegraphics[width=.95\linewidth]{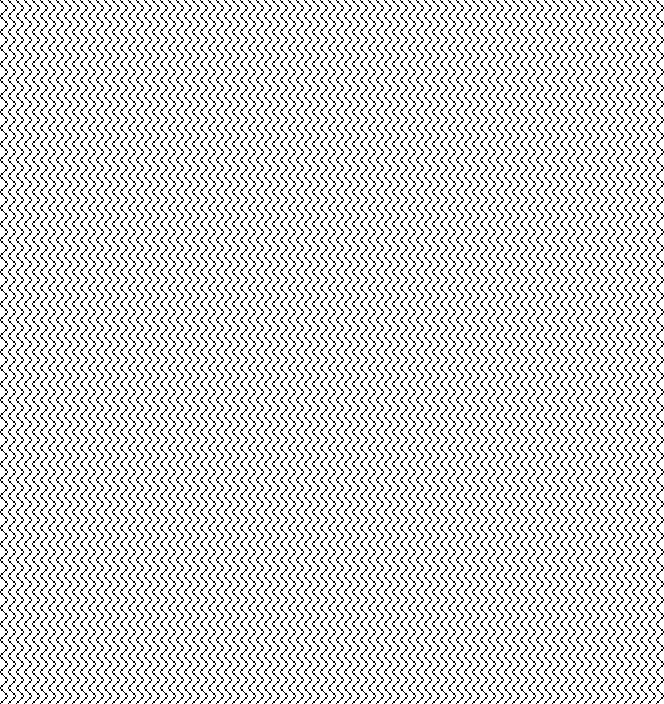}
        \adjincludegraphics[clip,trim=0 {.07\height} 0 {.15\height},width=.925\linewidth]{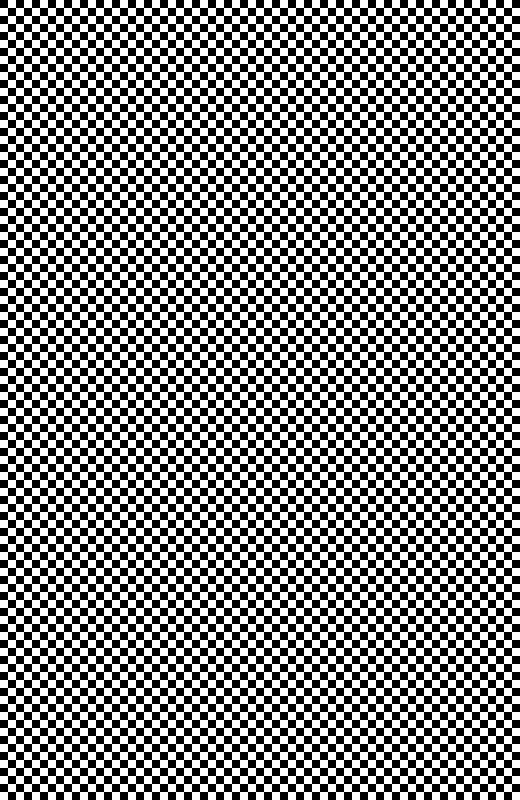}
    \end{minipage}
    }
    \subfloat[\cite{xie2020manga}]{
    \begin{minipage}[b]{.24\linewidth}
        \includegraphics[width=.95\linewidth]{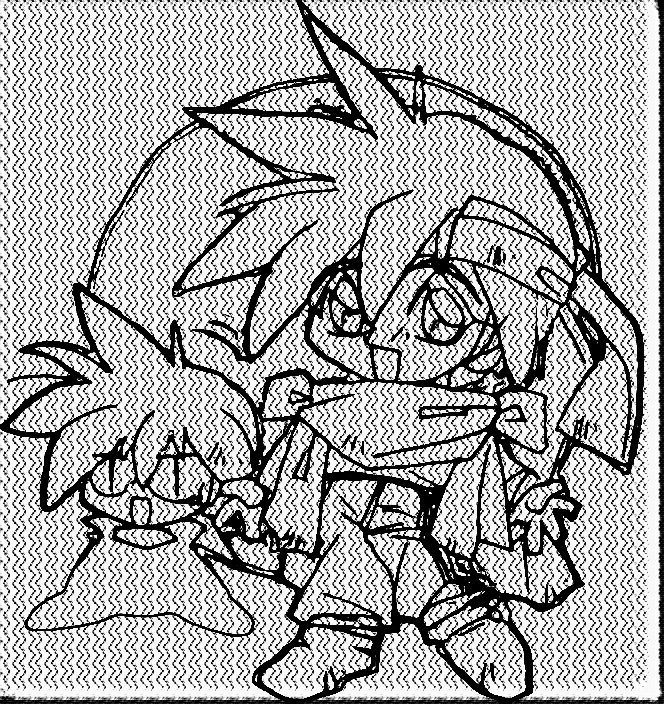}
        \adjincludegraphics[clip,trim=0 {.07\height} 0 {.15\height},width=.925\linewidth]{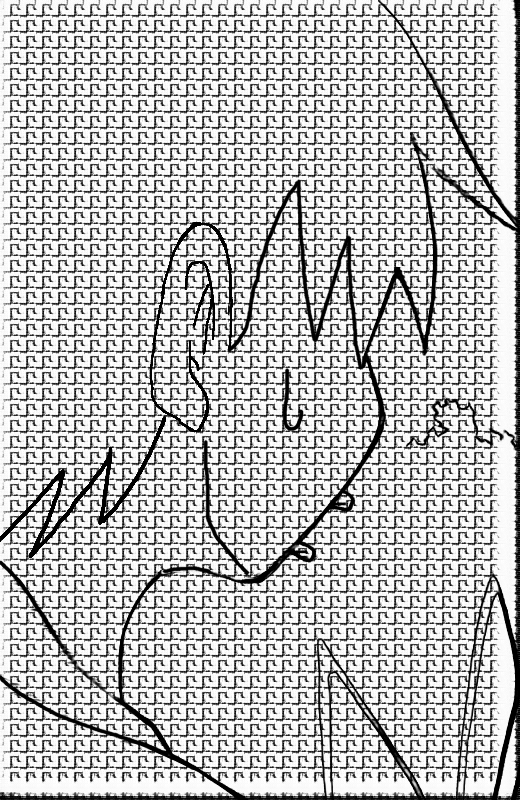}
    \end{minipage}
    }
    \subfloat[Ours]{
    \begin{minipage}[b]{.24\linewidth}
        \includegraphics[width=.95\linewidth]{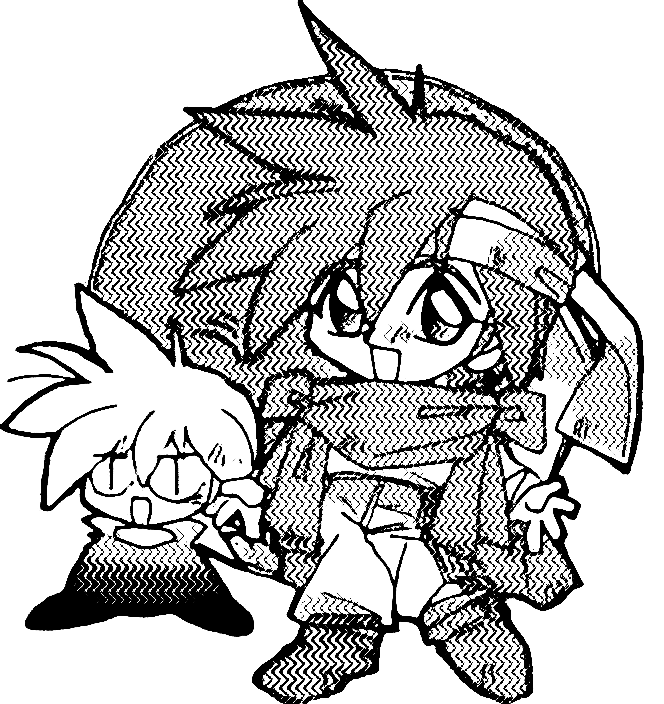}
        \adjincludegraphics[clip,trim=0 {.07\height} 0 {.15\height},width=.925\linewidth]{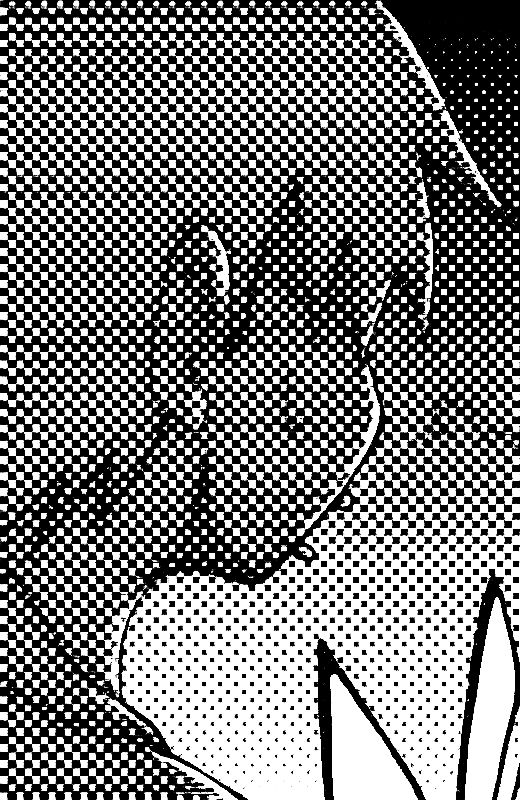}
    \end{minipage}
    }
    \vspace{-.07in}
    \caption{Our method can create smooth transitions of tone intensity while preserving the type of the screentone as well. ReveryEarth \copyright Yama Miyuki~\cite{mtap_matsui_2017}}
    \vspace{-.15in}
    \label{fig:interpolation}
\end{figure}

\textbf{Comparison on Controllable Manga Generation. }
Manga artists commonly use smoothly changing screentones to express shading or atmosphere, such as the hair of the third example in Fig.~\ref{fig:comparison} and the background in Fig.~\ref{fig:rescreening}, which requires the intensity feature to be interpolated in tone. With our design, our representation can provide controllable generation and manipulation on both the intensity and types of screentones. 
Fig.~\ref{fig:interpolation} (a) shows image synthesis by replacing the screentone type features of \cite{xie2020manga} and our method. Note that although~\cite{xie2020manga} can also generate various screentones, it is not providing stable controls of screentones, as the screentone intensity and type are non-linearly coupled in their model. As observed, the screentone types are disentangled with intensity in our model, and we can generate screentones with expected types while preserving the original intensity during rescreening.

\begin{figure*}[t!]
    \centering
    \subfloat[Screened manga]{
        \begin{minipage}[c]{.15\textwidth}
            \adjincludegraphics[clip,trim=0 0 0 {.25\width},width=\linewidth]{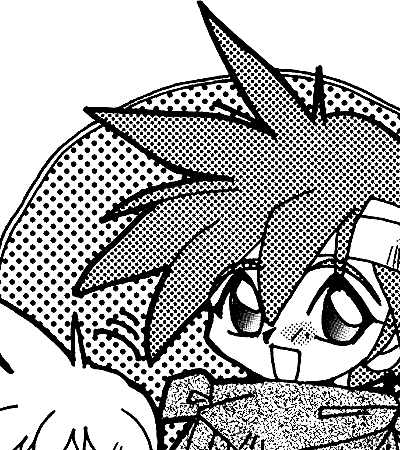}
        \end{minipage}}
    \subfloat[w/o $\mathcal{L}_{\rm adv}$]{
        \begin{minipage}[c]{.15\textwidth}
            \adjincludegraphics[clip,trim=0 0 0 {.25\width},width=\linewidth]{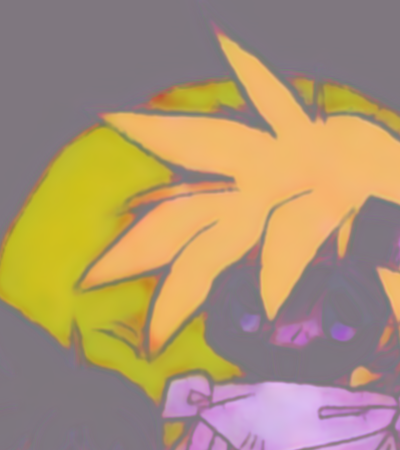}\\
            \adjincludegraphics[clip,trim=0 0 0 {.25\width},width=\linewidth]{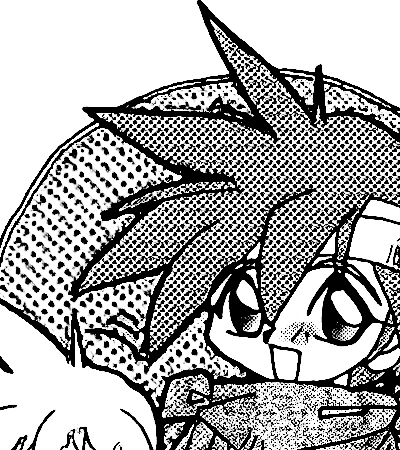}
        \end{minipage}}
    \subfloat[w/o $\mathcal{L}_{\rm kl}$]{
        \begin{minipage}[c]{.15\textwidth}
            \adjincludegraphics[clip,trim=0 0 0 {.25\width},width=\linewidth]{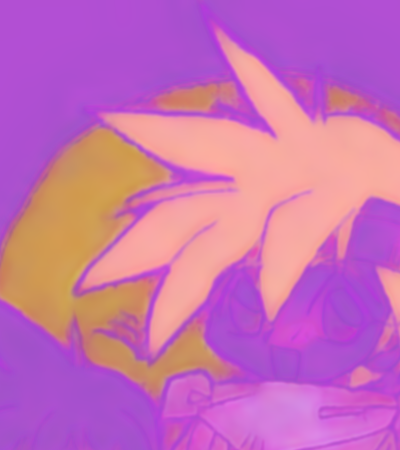}\\
            \adjincludegraphics[clip,trim=0 0 0 {.25\width},width=\linewidth]{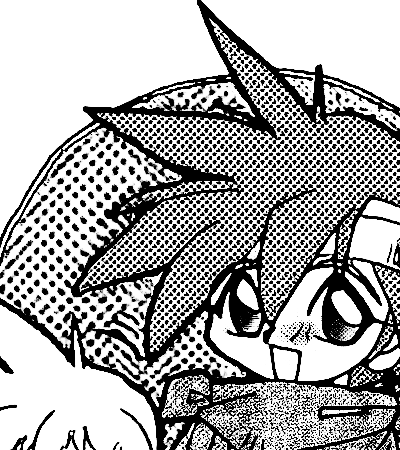}
        \end{minipage}}
    \subfloat[w/o $\mathcal{L}_{\rm fcons}$]{
        \begin{minipage}[c]{.15\textwidth}
            \adjincludegraphics[clip,trim=0 0 0 {.25\width},width=\linewidth]{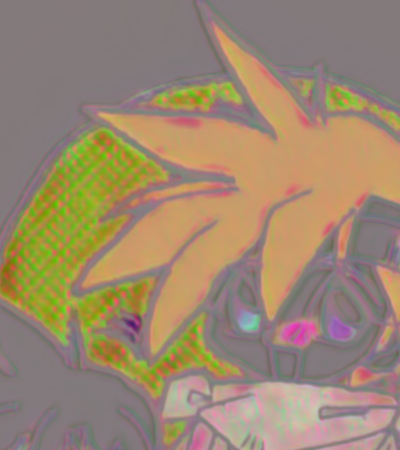}\\
            \adjincludegraphics[clip,trim=0 0 0 {.25\width},width=\linewidth]{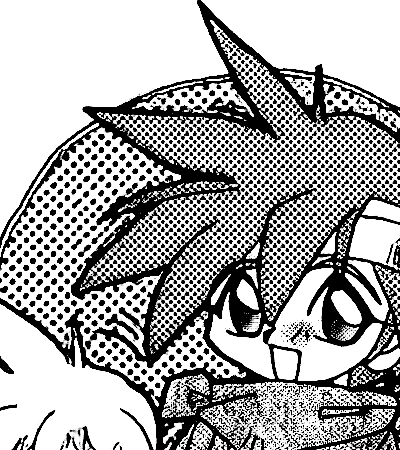}
        \end{minipage}}
    \subfloat[w/o $\mathcal{L}_{\rm frec}$]{
        \begin{minipage}[c]{.15\textwidth}
            \adjincludegraphics[clip,trim=0 0 0 {.25\width},width=\linewidth]{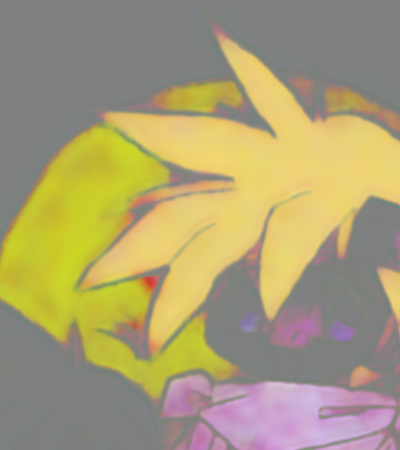}\\
            \adjincludegraphics[clip,trim=0 0 0 {.25\width},width=\linewidth]{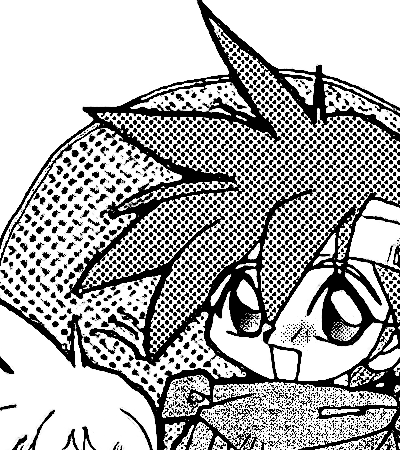}
        \end{minipage}}
    \subfloat[Ours]{
        \begin{minipage}[c]{.15\textwidth}
            \adjincludegraphics[clip,trim=0 0 0 {.25\width},width=\linewidth]{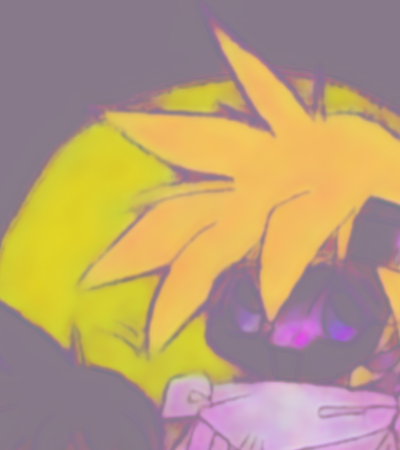}\\
            \adjincludegraphics[clip,trim=0 0 0 {.25\width},width=\linewidth]{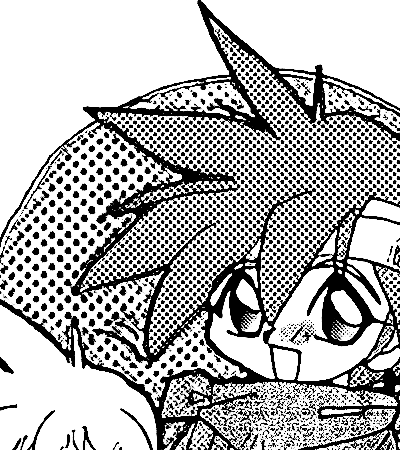}
        \end{minipage}}
    \vspace{-.05in}
    \caption{Impact of individual losses. From top to bottom, we show type feature and reconstructed image of each variant. ReveryEarth \copyright Yama Miyuki~\cite{mtap_matsui_2017}}
    \vspace{-.1in}
    \label{fig:ablationstudy}
\end{figure*}

\subsection{Quantitative Evaluation}
Besides visual comparison, we also quantitatively evaluate the performance of our interpretable representation. In this evaluation, we mainly compare with Gabor Wavelet~\cite{manjunath1996texture} and ScreenVAE~\cite{xie2020manga}. The evaluation is on four aspects: i) screentone summarization to measure the standard deviation within each screentone region; ii) screentone distinguishability to measure the standard deviation among different screentone regions; iii) intensity accuracy to measure the difference (Mean Absolute Error) between the generated intensity map and the target intensity map; iv) reconstruction accuracy to measure the difference (LPIPS~\cite{zhang2018perceptual}) between the input image and the reconstructed image. 
Table~\ref{tab:comparison} lists the evaluation results for all methods. Although Gabor Wavelet feature~\cite{manjunath1996texture} obtains the lowest value of summarization score, it got a low score on distinguishability, which means it is difficult to distinguish different screentones. ScreenVAE~\cite{xie2020manga} and ours achieve better performance for both screentone summarization and distinguishability. 
Furthermore, Gabor Wavelet feature~\cite{manjunath1996texture} cannot reconstruct the original screentones, while the other two methods can be used to synthesize the screentones. More importantly, our method explicitly extracts the intensity of manga, which is critical for practical manga parsing and editing.

\begin{table}[t!]
    \caption{Comparison with texture analysis methods.}
    \vspace{-.05in}
    \centering
    \begin{tabular}{p{0.15\textwidth}|p{0.07\textwidth}|p{0.07\textwidth}|p{0.07\textwidth}}
        \hline
        Property & \cite{manjunath1996texture} & \cite{xie2020manga} & Ours \\ \hline
        Summarization & \textbf{0.0141} & 0.0265  & 0.0207 \\ 
        Distinguishability & 0.0424 & 0.1715 & \textbf{0.2309} \\ \hline
        Intensity & - & - & \textbf{0.0262} \\ \hline
        Reconstruction & - & 0.3710 & \textbf{0.2434} \\ \hline
    \end{tabular}
    \label{tab:comparison}
\end{table}

\subsection{Ablation Study}
To study the effectiveness of individual loss terms, we performed ablation studies for each module by visually comparing the generated latent representation and the reconstructed image, as shown in Fig.~\ref{fig:ablationstudy}. Note that for better visualization, we did not normalize the type feature here. In addition, there is no loss configuration without reconstruction loss $\mathcal{L}_{\rm rec}$ or intensity loss $\mathcal{L}_{\rm itn}$. The reconstruction loss is necessary to preserve information in latent space, while our model will degrade to the original ScreenVAE~\cite{xie2020manga} without the intensity loss. 
Without adversarial loss $\mathcal{L}_{\rm adv}$, blurry and noisy screentones may be generated (background in the bottom row of Fig.~\ref{fig:ablationstudy}(b)). Without KL convergence loss $\mathcal{L}_{\rm kl}$, the latent space is not normally distributed and balanced representations may not be generated, which will fail the segmentation (top row of Fig.~\ref{fig:ablationstudy}(c)). Without feature consistency loss $\mathcal{L}_{\rm fcons}$, the network may not recognize multiscale screentones, and may generate inconsistent representation for the same screentone (top row of Fig.~\ref{fig:ablationstudy}(d)). Without feature reconstruction loss $\mathcal{L}_{\rm frec}$, inconsistent screentones may be generated among local regions (background in the bottom row of Fig.~\ref{fig:ablationstudy}(e)). 
In comparison, the combined loss can help the network recognize the type and intensity of screentones and generate a consistent appearance (Fig.~\ref{fig:ablationstudy}(f)).

\subsection{Limitations}
\begin{figure}[t!]
    \centering
    \subfloat[Input manga]{\includegraphics[width=.32\linewidth]{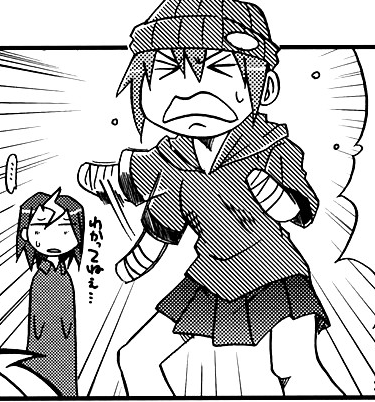}}\hspace{.01in}
    \subfloat[Type feature]{\includegraphics[width=.32\linewidth]{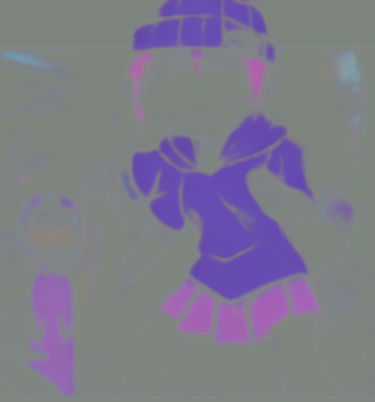}}\hspace{.01in}
    \subfloat[Segmentation]{\includegraphics[width=.32\linewidth]{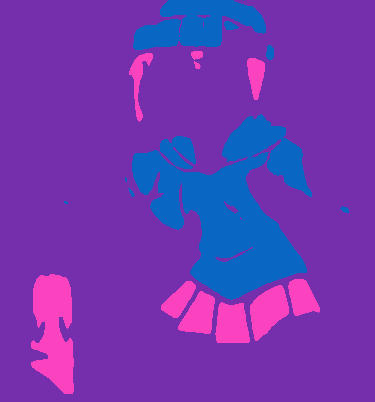}}
    \vspace{-.05in}
    \caption{Our method may fail to extract tiny screentone regions or recognize structure lines as strip screentones. }
    \vspace{-.05in}
    \label{fig:limitation}
\end{figure}

While our method can manipulate manga by segmenting the screentone region and generating screentones with desired screen type or intensity, our method may not extract some tiny screentones and some extra user hint may be required to generate good results (hair in Fig.~\ref{fig:limitation}(b)). In addition, some structural lines that are not extracted may be recognized as strip screentones (background in Fig.~\ref{fig:limitation}(b)). 

\section{Conclusion}
We propose an automatic yet controllable approach for rescreening regions in manga with user-expected screen types or intensity. It frees manga artists from the tedious manga rescreening process. 
In particular, we learn a screentone representation disentangled type and intensity of screentone, which is friendly for region discrimination and controllable screentone generation. With the interpretable representation, we can segment different screentone regions by measuring feature similarity. Users can generate controllable screentones to simulate various special effects.

{\small
\bibliographystyle{ieee_fullname}
\bibliography{reference}
}

\end{document}